\documentclass[sn-mathphys,Numbered]{sn-jnl}% Math and Physical Sciences Reference Style
\usepackage{graphicx}%
\usepackage{multirow}%
\usepackage{amsmath,amssymb,amsfonts}%
\usepackage{amsthm}%
\usepackage{mathrsfs}%
\usepackage[title]{appendix}%
\usepackage{xcolor}%
\usepackage{textcomp}%
\usepackage{manyfoot}%
\usepackage{booktabs}%
\usepackage{algorithm}%
\usepackage{algorithmicx}%
\usepackage{algpseudocode}%
\usepackage{listings}%

\usepackage[utf8]{inputenc}
\usepackage{amsmath}
\usepackage{amssymb}
\usepackage{enumitem}
\usepackage{pdfpages}
\usepackage{blkarray}
\usepackage{mathtools}
\usepackage{natbib}
\usepackage{url}
\usepackage{algorithm,algcompatible,amsmath}
\algnewcommand\INPUT{\item[\textbf{Input:}]}%
\algnewcommand\OUTPUT{\item[\textbf{Output:}]}%

\newcommand{\btheta}{\mbox{\boldmath{${\theta}$}}}

\def\bSigma{\mbox{\boldmath{$\Sigma$}}}
\def\mb{\mbox}
\def\smt{{\mbox{\tiny T}}}

\def\bX{\mathbf X}
\def\bx{{\bf x}}

\def\bz{{\bf z}}
\def\bZ{{\bf Z}}

\def\bB{\mathbf B}

\def\bW{\mathbf W}

\def\bX{\mathbf X}
\def\bx{\mathbf x}
\def\bI{\mathbf I}

\def\bR{\mathbf R}

\usepackage{color}
\usepackage{comment}
\usepackage{ulem}
\usepackage{soul}

\usepackage{physics}
\newcommand{\paren}{\pqty}

\newcommand{\mc}{\mathcal}

\usepackage[figuresright]{rotating}
\usepackage{graphicx}
\usepackage[capitalize,nameinlink]{cleveref}
\usepackage{setspace}

\def\bSig\mathbf{\Sigma}

\newcommand{\R}{\mathbb R}
\begin{document}

\title[iDeepViewLearn for Multiview Learning]{Interpretable Deep Learning Methods for Multiview Learning}

%%=============================================================%%
%% Prefix	-> \pfx{Dr}
%% GivenName	-> \fnm{Joergen W.}
%% Particle	-> \spfx{van der} -> surname prefix
%% FamilyName	-> \sur{Ploeg}
%% Suffix	-> \sfx{IV}
%% NatureName	-> \tanm{Poet Laureate} -> Title after name
%% Degrees	-> \dgr{MSc, PhD}
%% \author*[1,2]{\pfx{Dr} \fnm{Joergen W.} \spfx{van der} \sur{Ploeg} \sfx{IV} \tanm{Poet Laureate} 
%%                 \dgr{MSc, PhD}}\email{iauthor@gmail.com}
%%=============================================================%%

\author[1]{\fnm{Hengkang} \sur{Wang}}\email{wang9881@umn.edu}

\author[2]{\fnm{Han} \sur{Lu}}\email{lu000054@umn.edu}

\author[1]{\fnm{Ju} \sur{Sun}}\email{jusun@umn.edu}

\author*[2]{\fnm{Sandra E} \sur{Safo}}\email{ssafo@umn.edu}

\affil[1]{\orgdiv{Department of Computer Science and Engineering}, \orgname{University of Minnesota}, \orgaddress{\city{Minneapolis}, \postcode{55455}, \country{USA}}}

\affil*[2]{\orgdiv{Division of Biostatistics}, \orgname{University of Minnesota}, \orgaddress{\city{Minneapolis}, \postcode{55455}, \country{USA}}}

%%==================================%%
%% sample for unstructured abstract %%
%%==================================%%

\abstract{
\textbf{Background:} Technological advances have enabled the generation of unique and complementary types of data or views (e.g. genomics, proteomics, metabolomics) and opened up a new era in multiview learning research with the potential to lead to new biomedical discoveries.

\textbf{Results:} We propose iDeepViewLearn (Interpretable Deep Learning Method for Multiview Learning) to learn nonlinear relationships in data from multiple views while achieving feature selection. iDeepViewLearn combines deep learning flexibility with the statistical benefits of data and knowledge-driven feature selection, giving interpretable results. Deep neural networks are used to learn view-independent low-dimensional embedding through an optimization problem that minimizes the difference between observed and reconstructed data, while imposing a regularization penalty on the reconstructed data. The normalized Laplacian of a graph is used to model bilateral relationships between variables in each view, therefore, encouraging selection of related variables. iDeepViewLearn is tested on simulated and three real-world data for classification, clustering, and reconstruction tasks. For the classification tasks, iDeepViewLearn had competitive classification results with state-of-the-art methods in various settings. For the clustering task, we detected molecular clusters that differed in their 10-year survival rates for breast cancer. For the reconstruction task, we were able to reconstruct handwritten images using a few pixels while achieving competitive classification accuracy. The results of our real data application and simulations with small to moderate sample sizes suggest that iDeepViewLearn may be a useful method for small-sample-size problems compared to other deep learning methods for multiview learning.

\textbf{Conclusion:} iDeepViewLearn is an innovative deep learning model capable of capturing nonlinear relationships between data from multiple
views while achieving feature selection. It is fully open source and is freely available at \url{https://github.com/lasandrall/iDeepViewLearn}.


% \noindent\textbf{Availability:} Our algorithms are implemented in Pytorch and interfaced in R and available at:\\
% \url{https://github.com/lasandrall/iDeepViewLearn}.\\
\textbf{Contact:} {ssafo@umn.edu} or jusun@umn.edu\\
\textbf{Supplementary information:} Supplementary materials are available online.
}

\maketitle

\section{Background}
\label{s:intro}

Multiview learning has garnered considerable interest in biomedical research, thanks to advances in data collection and processing. Here, for the same individual, different sets of data or views (e.g., genomics, imaging) are collected, and the main interest lies in learning low-dimensional representation(s) common to all views or specific to each view that together explain the overall dependency structure among the different views.
%and can facilitate an understanding of the relationships among the views. 
Downstream analyses typically use the learned representations in supervised or unsupervised algorithms. For example, if a categorical outcome is available, then the learned low-dimensional representations could be used for classification. If no outcome is available, the low-dimensional representations could be used in clustering algorithms to cluster the samples. 
%Methods that jointly associate the views and predict an outcome  have also been proposed (e.g. \cite{SIDA:2019,BIPNet,palzer2021sjive}). In this paper, our primary focus is to develop methods to learn nonlinear low-dimensional representations in multiview data.  

\subsection{Existing Methods}
The literature on multiview learning is not scarce. Linear and nonlinear methods have been proposed to associate multiview data. For example, canonical correlation analysis (CCA) methods have been proposed to maximize the correlation between linear projections of two views \citep{Hotelling:1936, safo2018sparse}.  The kernel version of CCA (KCCA) has also been proposed to maximize the correlation between nonlinear functions of the views while restricting these nonlinear functions to reside in reproducing kernel Hilbert spaces \citep{Akaho:2001,lopez2014randomized}. 
%For large sample sizes, kernel methods can be computationally expensive as the time it requires to compute the representations scales poorly with the training sample size \citep{Andrew:2013,Kan:2016}. Randomized kernel CCA \citep{lopez2014randomized} have been proposed to alleviate the computational bottleneck for large sample sizes. 
Deep learning methods, which offer more flexibility than kernel methods, have been proposed to learn flexible nonlinear representations of two or more views, via deep neural networks (DNNs). Examples of such methods include Deep CCA \citep{Andrew:2013}, Deep generalized CCA [for three or more views] \citep{Benton2:2019} and DeepIMV \citep{lee2021variational}.
%and MOMA \cite{MOMA:2022}.
%Refer to \citep{} for a review of some nonlinear methods for associating data from multiple sources. 

Despite the success of DNN and kernel methods, their main limitation is that they do not yield interpretable findings.  In particular, if these methods are applied to our motivating data, it will be difficult to determine the genes and CpG sites that contribute the most to the dependency structure in the data. This is important for interpreting the results of downstream analysis that use these methods and for determining key molecules that discriminate between those who died from breast cancer and those who did not. 

Few interpretable deep-leaning methods for multiview learning have been proposed in the literature. In \cite{MOMA:2022}, a data integration and classification method (MOMA) was proposed for multiview learning that uses the attention mechanism for interpretability. Specifically, MOMA builds a module (e.g. gene set) for each view and uses the attention mechanism to identify modules and features relevant to a certain task. 

In \cite{wang2021deep}, a deep learning method was proposed to jointly associate data from multiple views and discriminate subjects that allows for feature ranking. The authors considered a homogeneous ensemble approach for feature selection that allowed the ranking of features based on their contributions to the overall dependency among views and the separation of classes within a view. It is noteworthy that variable selection in MOMA and Deep IDA is data driven, and the algorithm for MOMA is applicable to two views, which is very restrictive.

\subsection{Our Approach}
In this article, we propose a deep learning framework to associate data from two or more views while achieving feature selection. Similar to deep generalized CCA [deep GCCA] \citep{Benton2:2019} and unlike deep CCA \citep{Andrew:2013}, we learn low-dimensional representations that are common to all views. However, unlike deep GCCA, we assume that each view can be approximated by a nonlinear function of the shared low-dimensional representations. We use deep neural networks to model the nonlinear function and construct an optimization problem that minimizes the difference between the observed and the nonlinearly approximated data, while imposing a regularization penalty on the reconstructed data. This allows us to reconstruct each view using only the relevant variables in each view. As a result, the proposed method allows the selection of variables in the views and enhances our ability to identify features from each view that contribute to the association of the views. The results of our motivating data and simulations with small sample sizes suggest that the proposed method may be a useful method for small-sample-size problems compared to other deep learning methods for associating multiple views. Beyond the data-driven approach to feature selection, we also consider a knowledge-based approach to identify relevant features. In the statistical learning literature, the use of prior information (e.g., biological information in the form of variable-variable interactions) in variable selection methods has the potential to identify correlated variables with greater ability to produce interpretable results and improve prediction or classification estimates \citep{SIDA:2019,safo2018integrative}. As such, we use the normalized Laplacian of a graph to model bilateral relationships between variables in each view and to encourage the selection of variables that are connected.   

In summary, we have three main contributions. First, we propose a deep learning method for learning nonlinear relationships in multiview data that is capable of identifying relevant features that contribute the most to the association among different views. Our approach can accommodate more than two views, in contrast to MOMA, which requires significant code modifications by users, for the same purpose. Second, we extend this method to incorporate prior biological information to yield more interpretable findings, distinguishing it from existing interpretable deep learning methods for multiview learning, such as MOMA and Deep IDA. To the best of our knowledge, this is one of the first nonlinear-based methods for multiview learning to do so. Third, we provide an efficient implementation of the proposed methods in Pytorch and interface them in R to increase the reach of our algorithm. 

The remainder of the paper is organized as follows. In \cref{s:method}, we introduce the proposed method. In \cref{s: sim}, we conduct simulation studies to assess the performance of our methods compared to several existing linear and nonlinear methods. In \cref{s: real}, we apply our method to the Holm breast cancer study for classification and clustering; we further consider two additional applications: brain lower grade glioma (LGG) data, to demonstrate the use of our method for three views; and MNIST handwriting data, to demonstrate that handwritten digits can be reconstructed with few pixels while maintaining competitive classification accuracy.

% ==================== METHOD ====================

\section{Methods}
\label{s:method}

\subsection{Model Formulation}

Assume that $d=1,\ldots,D$ different types of data or views are available from $n$ individuals and organized in $D$ matrices $\mathbf{X}^{(1)} \in \R^{n \times p^{(1)}}$, $\dots$, $\mathbf{X}^{(D)} \in \R^{n \times p^{(D)}}$. For example, for the same set of $n$ individuals in our motivating study, the matrix $\mathcal{\mathbf{X}}^{(1)}$ consists of gene expression levels and $\mathcal{\mathbf{X}}^{(2)}$ consists of CpG sites. Denote an outcome variable by $\mathbf{y}$, if available.  In our motivating study, $\mathbf{y}$ is an indicator variable of whether or not an individual died from breast cancer. We wish to model complex nonlinear relationships between these views via an \textit{informative} joint low-dimensional nonlinear embedding of the original high-dimensional data. 
%We have two main goals. First, we wish to model complex nonlinear relationships among these different views via a joint low-dimensional nonlinear embedding of the original high-dimensional data. Second, we wish to identify the important variables that contribute the most to the nonlinear relationships among the views. %Once the low

For the sake of clarity, we outline a linear framework which our nonlinear model emulates. Assume that there is a joint embedding (or common factors) $\mathbf{Z} \in \R^{n \times K}$ of the $D$ views that drives the observed variation across the views so that each view is written as a linear function of the joint embedding plus some noise: $\bX^{(d)} = \mathbf{Z} \mathbf{B}{^{(d)}}^{\smt} + \mathbf E^{(d)}.$
Here, $K$ is the number of latent components and $\mathbf{B}{^{(d)}} \in \R^{p^{(d)} \times K}$ is the loading matrix for view $d$, each row corresponding to the coefficients $K$ for a specific variable. $\mathbf E^{(d)}$ is a matrix of errors incurred by approximating $\bX^{(d)}$ with $\mathbf{Z} \mathbf{B}{^{(d)}}^{\smt}$. Let $\bz_i \in \R^K$ be the $i$th row in $\bZ$. The common factors $\bz_i$ represent $K$ different driving factors that predict all variables in all views for the $i$th subject, thus inducing correlations between views. 
%Equation (\ref{eqn:main}) assumes that each view  can be approximated by $\mathbf{Z}\mathbf B{^{(d)}}^{\smt}$. 
%\emph{Second}, we assume that the outcome $\bf{y}$ depends on the $d$th data source $\bf X^{d}$ through the common factor $\bm Z$; this allows us to couple integrative analysis with the prediction of a clinical outcome simultaneously. \\
%\subsection{Nonlinear Simulations}
When we write $\bX^{(d)} \approx \bZ \mathbf{B}{^{(d)}}^{\smt}$ for $d = 1, \dots, D$, we assume that there is an ``intrinsic" space $\R^K$ so that each sample is represented as $\bz \in \R^K$. For each $d = 1, \dots, D$, $\bx^{(d)}$ is an instance in $\bX^{(d)}$, and $\mathbf{B}^{(d)}$ maps a low-dimensional representation $\bz$ to this $\bx^{(d)}$, i.e., restricting the mappings to be linear. Now, we generalize these mappings to be nonlinear, parameterized by neural networks. 
 
For $d = 1, \dots, D$, let $G_d$ denote the neural network that generalizes $\bB^{(d)\smt}$ for the view $d$. As typical neural networks, each of the $G_d$'s is composed of multilayer affine mapping followed by nonlinear activation, i.e., of the form 
% \begin{align}
    $\mc W_L \circ \sigma \circ \mc W_{L-1} \dots \sigma \circ \mc W_2 \circ \sigma \circ \mc W_1$,  
% \end{align} 
where $\sigma$ denotes the nonlinear activation applied element-wise, and $\mc W_i$' s for $i = 1, \dots, L$ denote the affine mappings. We prefer to state the affine layers in abstract form, as we can have different types of layer. In this paper, we use $G_d$ consisting of fully-connected and convolutional layers~\cite{wang_early_2021,chen_detecting_2023} to reconstruct numerical data and images, respectively.

For simplicity, assume that each layer of the $d$th view network, except the first layer, has $c_d$ units.  Let the size of the input layer (first layer) be $K$, where $K$ is the number of latent components. The output of the first layer for the $d$th view is a function of the shared low-dimensional representation, $\bZ$, and is given by $h_1^{(d)}= \sigma (\mathbf{Z}\mathbf{W}_1^{(d)} + \mathbf{b}_1^{(d)}) \in \R^{n \times c_d} $ where $\mathbf{W}_1^{(d)} \in \R^{K \times c_d} $ is a matrix of weights for view $d$, $\mathbf{b}_1^{(d)} \in \R^{n \times c_d}$ is a matrix of biases, and $\sigma :\R \longrightarrow \R$ is a nonlinear mapping. The output of the second layer for the $d$th view is the $h_2^{(d)}=\sigma (h_1^{(d)}\mathbf{W}_2^{(d)} + \mathbf{b}_2^{(d)}  ) \in \R^{n \times c_d}$, $\mathbf{W}_2^{(d)} \in \R^{K \times c_d}$ matrix of weights, $\mathbf{b}_2^{(d)} \in \R^{n \times c_d}$  matrix of biases. The final output layer for the $d$th view is given by $G_d(\bZ)=\sigma (h_{(K_d-1)}^{(d)} \mathbf{W}_{K_d}^{(d)} + \mathbf{b}_{K_d}^{(d)}  ) \in \R^{n \times p^{(d)}}$, $h_{(K_d-1)}^{(d)} \in \R^{n \times c_d}$, $\mathbf{W}_{K_d}^{(d)} \in \R^{c_d \times p^{(d)}}$,  $\mathbf{b}_{K_d}^{(d)}   \in \R^{n \times p^{(d)}}$, and the subscript $K_d$ denotes the $K$th hidden layer for the view $d$. $G_d(\bZ)$ is a function of the weights and biases of the network. 

Our first goal is to approximate each view with a nonlinear embedding of the joint low-dimensional representation {in an interpretable manner}, i.e., $\bX^{(d)} \approx G_d(\bZ)$. {To achieve interpretability, MOMA used the attention mechanism to choose important features. In the statistical learning literature, regularization techniques (e.g.,  lasso \cite{lasso}, elastic net \cite{enet}, SCAD \cite{SCAD}) are oftentimes used for variable selection to promote interpretability. We also propose a regularization approach for interpretability.} Specifically, we assume that some variables in $\bX^{(d)}$ are irrelevant and are not needed in the approximation of $\bX^{(d)}$. Thus, the columns of $G_d(\bZ)$ corresponding to the unimportant variables in $\bX^{(d)}$ should be made zero or nearly zero in the nonlinear approximation of $\bX^{(d)}$. To achieve this, we adopt the $\ell_{2,1}$ norm from~\cite{DBLP:conf/nips/NieHCD10} to promote column-wise sparsity for features, where $\ell_{2,1}$ is denoted as follows:
%\begin{align}\label{eqn:l21}
$\norm{\bf X}_{2, 1} = \sum_{j=1}^p \sqrt{\sum_{i=1}^n \mb X_{ij}},~ \bf X \in \R^{n \times p}$, where $X_{ij}$ is the $ij$th element in $\bX$.
%\end{align}
Given these assumptions, we propose to solve the following optimization problem: find the parameters of the neural network (weight matrices, biases) defining the neural network $G_d$, and the shared low-dimensional representation $\bZ$, for $d=1,\ldots,D$ that 
\begin{align}\label{eqn:nonlinear}
 \min_{ \mathbf{W}^{(1)}, \cdots, \mathbf{W}^{(D)}, \mathbf{b}^{(1)},\cdots,\mathbf{b}^{(D)}, \bZ} \;  \sum_{d=1}^D \paren{\|\bX^{(d)} - G_{d}\paren{\bZ}\|_{2, 1} + \lambda^d\|G_{d}\paren{\bZ}\|_{2, 1}}.
\end{align}
% Here, $\norm{\cdot}_{2, 1}$ helps to promote columnwise sparsity and, hence, selection of important features across samples. Thus, t
The two terms $\|\bX^{(d)} - G_{d}\paren{\bZ}\|_{2, 1} + \lambda^d \|G_{d}\paren{\bZ}\|_{2, 1}$ together ensure that we select a subset of columns from $\bX^{(d)}$ to approximate $\bX^{(d)}$. $\lambda^d$'s are regularization parameters that could be selected by $k$-fold cross-validation, where $k = 5$ throughout this paper. 
%We found $\lambda^d$ to be insensitive, so we fix these to $0.1$ for each view, to improve computational efficiency.

Although $\norm{\cdot}_{2, 1}$ helps promote column-wise sparsity, we did observe that the columns of $G_{d}\paren{\bZ}$ were not exactly zero across all samples but were shrunk towards zero for noise variables, perhaps as a result of our use of an automatic differentiation function. Thus, we proceed as follows to select/rank features.  Once we have learned the latent code $\bZ$ and neural networks $G_1$, $G_2$, ..., $G_D$, we use this information to obtain reconstructed data for the different views, that is, to obtain $G_d(\bZ)$ for the view $d$. We then calculate the column-wise $l_2$ norm of $G_d(\bZ)$, and choose the top $r$\% columns with the largest column norms as important features for the corresponding view. It is imperative that the variables in each view be on the same scale in order to use this ranking procedure. Thus, we standardize each variable to have mean zero and variance one.  We save the indices of important features as $\bI_1$, $\bI_2$, ...,$\bI_D$, and we denote the new datasets with the selected indices as $\bX^{\prime (1)}, \bX^{\prime (2)}, ..., \bX^{\prime (D)}$. 

Compared to existing deep learning methods for associating multiple views (e.g., deep generalized CCA), our formulation (\ref{eqn:nonlinear}) is unique because we learn the shared low-dimensional representation, $\bZ$, while also selecting important variables in each view that drive the association among views. Similarly to Deep GCCA, and unlike CCA and Deep CCA that only learn linear transformations of each view, we learn $\bZ$, which is independent of the views and allows one to reconstruct all the view-specific representations simultaneously. \cref{fig:fs} is a schematic representation to train the neural network and select features. After that, downstream analyses can use the learned $\bZ$ in classification, regression, and clustering algorithms, as shown in \cref{fig:train}.

\subsection{Network-based feature selection}

\begin{figure}[!htbp]
    \centering
    \includegraphics[width=0.9\linewidth]{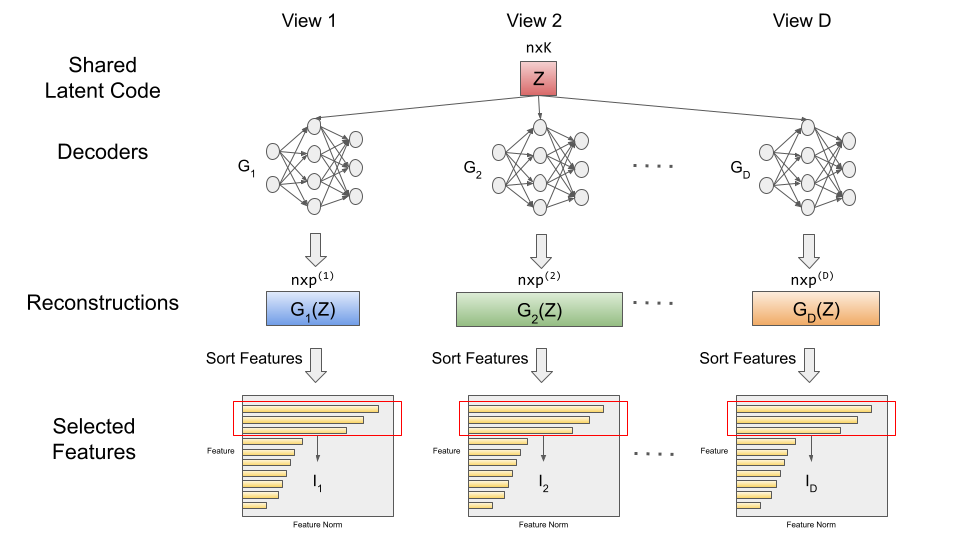}
    \caption{Feature Selection. We train a deep learning model that takes all the views, estimates a shared low-dimensional representation $\bZ$ that drives the variation across the views, and obtains nonlinear reconstructions ($G_1(\bZ)$,\ldots,$G_D(\bZ)$) of the original views. We impose sparsity constraints on the reconstructions allowing us to identify a subset of variables for each view ($\bI_1$, \ldots,$\bI_D$) that  approximate the original data.}
    \label{fig:fs}
\end{figure}

We consider a knowledge-based approach to identify potentially relevant variables that drive the dependency structure among views (see \cref{fig:fs}). %We assume that the important variables in $\bX^{(d}$ are connected together, as such using information about variable-variable interactions can help 
In particular, we use prior knowledge about variable-variable interactions (e.g., protein-protein interactions) in the estimation of $G_d\paren{\bZ}$. Incorporating prior knowledge about variable-variable interactions can capture complex bilateral relationships between variables. It has the potential to identify functionally meaningful variables (or networks of variables) within each view for improved prediction performance, as well as aid in interpretation of variables. 

There are many databases for obtaining information on variable-variable relationships. One of such database for protein-protein interactions is the Human Protein Reference Database (HPRD) \citep{hprd2003}. 
We capture the variable-variable connectivity within each view in our deep learning model using the normalized Laplacian \citep{chung1997spectral} obtained from the graph underlying the observed data. Let $\mathcal{G}^{(d)} =(V^{(d)},E^{(d)},W^{(d)})$, $d=1,2,\ldots,D$ be a graph network given by a weighted undirected graph. $V^{(d)}$ is the set of vertices corresponding to the $p^{(d)}$ variables (or nodes) for the $d$-th view. Let $E^{(d)}=\{u \sim v\}$ if there is an edge of variable $u$ to $v$ in the $d$th view. $W^{(d)}$ is the weight of an edge for the $d$-th view that satisfies $w(u,v) =w(v,u) \ge 0$. %  Note that if $\{u,v\} \not \in E(\mathcal{G})$, then $w(u,v)=0$. 
Denote $r_v$ as the degree of vertex $v$ within each view; $r_v = \sum_{u}w(u,v)$.
The normalized Laplacian of $\mathcal{G}^{(d)}$ for the $d$-th view is
$ \mathcal{L}^{(d)}= T^{-1/2}LT^{-1/2}$ where $L$ is the Laplacian of a graph defined as
%\begin{small}
\begin{eqnarray} \label{eqn:laplacian}
L(u,v)  &=& \begin{cases}
r_v-w(u,v) & \mbox{if ~$u$~$=$~$v$}  \\
-w(u,v) & \mbox{if ~$u$~ and ~$v$~ are~adjacent} 
\\
0 &~ \mbox{otherwise},\ 
\end{cases} 
\end{eqnarray}
 and $T$ is a diagonal matrix with $r_v$ as the (u,v)-th entry. 
%\end{small}
%The matrix $\mathcal{L}_n(u,v)$ is usually sparse (has many zeros) and so can be stored with sparse functions in any major software programs such as R or Matlab.
Given $\mathcal{L}^{(d)}$, we solve the problem: 
\begin{align}\label{eqn:nonlinearsmooth}
\min_{ \mathbf{W}^{(1)}, \cdots, \mathbf{W}^{(D)}, \mathbf{b}^{(1)},\cdots,\mathbf{b}^{(D)}, \bZ}  \;  \sum_{d=1}^D \paren{\|\bX^{(d)} - G_d\paren{\bZ}\|_{2, 1} + \lambda^d\|G_d\paren{\bZ}\mathcal{L}^{(d)}\|_{2, 1}}.
\end{align}
The normalized Laplacian $\mathcal{L}^{(d)}$ is used as a smoothing operator to smooth the columns in $G_d\paren{\bZ}$ so that the variables connected in the $d$-th view are encouraged to be selected together.

\subsection{Prediction of shared low-dimensional representation and downstream analyses}

\begin{figure}[!htbp]
    \centering
    \includegraphics[width=0.9\linewidth]{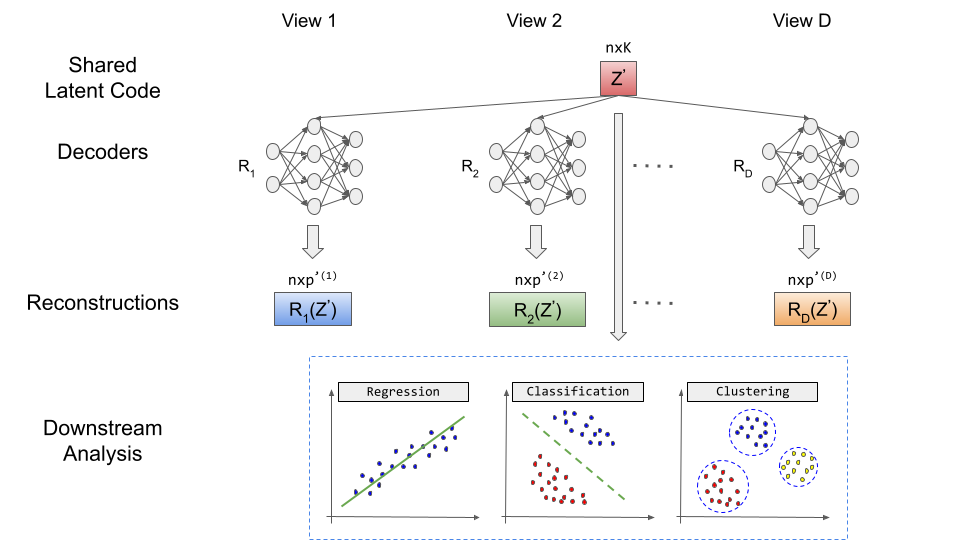}
    \caption{Reconstruction and Downstream Analysis. We train a deep learning model to obtain a common low-dimensional representation $\bZ'$ that is based on the features selected in Algorithm 1, we obtain nonlinear approximations ($\bR_1(\bZ)$,\ldots,$\bR_D(\bZ))$, and we perform downstream analyses using estimated $\bZ'$.}
    \label{fig:train}
\end{figure}

In this section, we would like to predict the low-dimensional representation shared from the test data $\bX^{(d)}_{test}, d=1,\cdots,D$, (i.e., $\bZ_{test}$) and use this information to predict an outcome, $\mathbf{y}$, if available. The schematic graph is shown in \cref{fig:train}. Note that $\mathbf{y}$ can be continuous, binary, or multiclass. We discuss our approach to predict the shared low-dimensional representation, $\bZ_{test}$. After getting important features of $\bX^{(d)}$ using equation (\ref{eqn:nonlinear}) or (\ref{eqn:nonlinearsmooth}) , we extract these features from the original training dataset and form a new training dataset ${\bX}^{'(d)}$. We also form a new testing dataset $\bX^{'(d)}_{test}  d=1,\cdots, D$ that consists of the important features. Let $p^{'(d)}$  denote the cardinality of the columns in view $d$. Since the $\bZ$ learned in Equation (\ref{eqn:nonlinear}) or (\ref{eqn:nonlinearsmooth}) is estimated using important and unimportant features, when used in downstream analyses, it can lead to poor results. Therefore, we construct a new shared low-dimensional representation, ${\bZ'}$, which is based only on important features, that is, ${\bX}^{'(d)}, d=1,\ldots,D$. 
%Specifically,  we start another round of reconstruction process to learn a ${\bZ'}$ of the new dataset ${\bX}^{'(d)}$ instead of directly utilizing the learned $\bZ$ in equation (\ref{eqn:nonlinear}) or (\ref{eqn:nonlinearsmooth}). We take this approach because $\bZ$ was estimated using both important and unimportant features. 
Because we have already selected a subset of relevant columns of $\bX^d$, we are willing to have non-sparse reconstruction results. Therefore, we find ${\bZ'}$ by solving the optimization problem:
\begin{small}
\begin{align}\label{eqn:nonlinear2}
% (\widetilde{\bW}^{'(1)},\ldots,\widetilde{\bW}^{'(D)}, \widetilde{\mathbf{b}}^{'(1)},\ldots,\widetilde{\mathbf{b}}^{'(D)}, \widetilde{\bZ'} )= \min_{ \mathbf{W}^{'(1)}, \cdots, \mathbf{W}^{'(D)}, \mathbf{b}^{'(1)},\cdots,\mathbf{b}^{'(D)}, \bZ'} \;  \sum_{d=1}^D \|{\bX}^{'(d)} - {R}_d\paren{\bZ'}\|_{F}^2,
\min_{ \mathbf{W}^{'(1)}, \cdots, \mathbf{W}^{'(D)}, \mathbf{b}^{'(1)},\cdots,\mathbf{b}^{'(D)}, \bZ'} \;  \sum_{d=1}^D \|{\bX}^{'(d)} - {R}_d\paren{\bZ'}\|_{F}^2,
\end{align}
\end{small}
where $R_d$ depends on the weights of the network parameters $\bW^{'(d)}$ and the biases $\mathbf{b}^{'(d)}$, and $\|\cdot\|_F$ is the Frobenius norm. Specifically, the final output $R_d= \sigma(h_{(K_d-1)}^{(d)} {\bW}_{K_d}^{'(d)} + {\mathbf{b}}_{K_d}^{'(d)}) \in  \R^{n \times p^{'(d)}}$, the subscript $K_d$ denotes the $K$th hidden layer for view $d$, and the first hidden layer is given as $h_1=\sigma(\mathbf{Z'}{\bW}_1^{'(d)} + {\mathbf{b}}_1^{'(d)})$.
%Denote the nueral networks learned (weights and biases) by $\mathbf{R}_1,\cdots,\mathbf{R}_D$. 

Suppose that $\widetilde{{R}_d}\paren{\widetilde{\bZ'}}$ can approximate ${\bX}^{'(d)}$ well for each view, that is, ${\bX}^{'(d)} \approx \widetilde{{R}_d}\paren{\widetilde{\bZ'}}, d=1,\cdots,D$. Then it is easy to find ${\bX}^{'(d)} \approx \paren{\tau \widetilde{{R}_d}}\paren{\widetilde{\frac{\bZ'}{\tau}}}$ for any $\tau \in \R_{\ne 0}$ because $\widetilde{{R}_d}$ and $\widetilde{\bZ'}$ are optimized simultaneously. The lack of control of the scaling of the learned representation $\widetilde{\bZ'}$ can lead to robustness problems in downstream analysis, so we add additional constraints on $\bZ'$ in equation (\ref{eqn:nonlinear2}). However, since it is likely that the shape of $\bZ'$ is not the same as the latent code learned in the testing stage due to the different number of samples, we put constraints on each row of $\bZ'$ (we assume that the number of latent components in the training and testing data is the same) as $\| \bz'_{i} \|_2 \leq 1$ where $\bz'_{i}$ means the $i$-th row vector of $\bZ'$, that is, the latent code of the $i$th sample. Finally, the optimization problem is as follows:

\begin{small}
\begin{multline}\label{eqn:nonlinear2_new}
% (\widetilde{\bW}^{'(1)},\ldots,\widetilde{\bW}^{'(D)}, \widetilde{\mathbf{b}}^{'(1)},\ldots,\widetilde{\mathbf{b}}^{'(D)}, \widetilde{\bZ'} )= \min_{ \mathbf{W}^{'(1)}, \cdots, \mathbf{W}^{'(D)}, \mathbf{b}^{'(1)},\cdots,\mathbf{b}^{'(D)}, \bZ'} \;  \sum_{d=1}^D \|{\bX}^{'(d)} - {R}_d\paren{\bZ'}\|_{F}^2,
\min_{ \mathbf{W}^{'(1)}, \cdots, \mathbf{W}^{'(D)}, \mathbf{b}^{'(1)},\cdots,\mathbf{b}^{'(D)}, \bZ'} \;  \sum_{d=1}^D \|{\bX}^{'(d)} - {R}_d\paren{\bZ'}\|_{F}^2,
\textrm{s.t.} \| \bz'_{i} \|_2 = 1, i=1,\ldots,n
\end{multline}
\end{small}

To learn $\bZ'_{test}$ from the test data $\bX^{'(d)}_{test}, d=1,\cdots,D$, we use the weights of the learned neural network, $\widetilde{\bW}^{'(1)},\ldots,\widetilde{\bW}^{'(D)}$ and biases $\widetilde{\mathbf{b}}^{'(1)},\ldots,\widetilde{\mathbf{b}}^{'(D)}$ and we solve the following optimization problem for $\widetilde{\bZ}_{test}$:
\begin{align}\label{eqn:nonlinear3}
% \widetilde{\bZ'}_{test}= \min_{\bZ'_{test}} \;  \sum_{d=1}^D \|\bX^{'(d)}_{test} - \widetilde{R}_{d}\paren{\bZ'_{test}}\|_{F}^2,
% \; \textrm{s.t.} \| \bz'_{test_i} \|_2 = 1, i=1,\ldots,n.
\min_{\bZ'_{test}} \;  \sum_{d=1}^D \|\bX^{'(d)}_{test} - \widetilde{R}_{d}\paren{\bZ'_{test}}\|_{F}^2,
\; \textrm{s.t.} \| \bz'_{test_i} \|_2 = 1, i=1,\ldots,n.
\end{align}
Here,  $\bz'_{test_i}$ is the $i$th row vector in $\bZ_{test}$ and $\bX^{'(d)}_{test}$ refers to the testing dataset with column indices $\bI_d$, i.e.,, only the columns that are selected as important are used to estimate $\widetilde{\bZ'}_{test}$. 
%Denote the cardinality of the columns in view $d$ as $p^{'(d)}$.   
The output layer $\widetilde{R}_{d}=
\sigma(h_{(K_d-1)}^{(d)} \widetilde{\bW}_{K_d}^{'(d)} + \widetilde{\mathbf{b}}_{K_d}^{'(d)}) \in \R^{n \times p^{'(d)}}$, the subscript $K_d$ denotes the $K$th hidden layer for view $d$,  $h_{(K_d-1)}^{(d)} \in \R^{n \times c_d}$, $\widetilde{\bW}_{K_d}^{'(d)} \in \R^{c_d \times p^{'(d)}}$, and $\mathbf{b}_{K_d}^{'(d)}   \in \R^{n \times p^{(d)}}$, and $h_1=\sigma(\mathbf{Z}'_{test}\widetilde{\bW}_1^{'(d)} + \widetilde{\mathbf{b}}_1^{'(d)})$.

Now, when predicting an outcome, the low-dimensional representations $\widetilde{\bZ'}$ and $\widetilde{\bZ'}_{test}$ become training and testing data, respectively. For example, to predict a binary or multiclass outcome, we train a support vector machine (SVM)~\citep{ben2008support} classifier with the training data $\widetilde{\bZ'}$  and the outcome data $\mathbf{y}$, and we use the learned SVM model and the testing data $\widetilde{\bZ'}_{test}$ to obtain the predicted class membership,  $\widehat{\mathbf{y}}_{test}$. We compare $\widehat{\mathbf{y}}_{test}$ with  $\mathbf{y}_{test}$ and we estimate the classification accuracy. For continuous outcome, one can implement a nonlinear regression model and then compare the predicted and true outcomes using a metric such as the mean squared error (MSE). For unsupervised analyses, such as clustering, an existing clustering algorithm, such as K-means clustering, can be trained on $\widetilde{\bZ'}$. \cref{fig:train} is a schematic representation of the prediction algorithm and the downstream analyses proposed.

We provide our optimization approach in the Supplementary Material.  Our algorithm is divided into three stages. The first stage is the Feature Selection stage. In this stage, we solve the optimization problem (\ref{eqn:nonlinear}) or (\ref{eqn:nonlinearsmooth}) to obtain features that are highly ranked. The second stage is the Reconstruction and Training stage using selected features. Here, we solve the optimization problem (\ref{eqn:nonlinear2}). Our input data are the observed data with the selected features (that is, the top $r$ or $r\%$ features in each view), $\bX^{'(1)}\ldots \bX^{'(D)}$.  At convergence, we obtain the reconstructed data $R_d(\bZ')$, and the learned shared low-dimensional representations $\widetilde{\bZ'}$ based only on the top $r$ or $r\%$ variables in each view. Downstream analyses such as classification, regression, or clustering could be carried out on these shared low-dimensional representations learned. The third stage is the prediction stage, if an outcome is available. Here, we solve the optimization problem (\ref{eqn:nonlinear3}) for the learned shared low-dimensional representation ($\widetilde{\bZ}'_{test}$) corresponding to the test views ($\bX^{'(1)}_{test} \ldots \bX^{'(D)}_{test}$). This can be used to obtain prediction estimates (e.g. testing classification via an SVM model).

% ==================== SIMULATION ====================

\section{{Simulation Experiments}}
\label{s: sim}
We conducted simulation studies to assess the performance of iDeepViewLearn for varying data dimensions,
as the relationship between views becomes more complex and when prior information on variable-variable relationships is available or not. Please refer to the Supplementary Materials for more simulation setup and results.

\subsection{Set-up when there is no prior information on variable-variable interactions}
%We demonstrate the classification performance when the low-dimensional representations are used in a classification algorithm.  
We consider two different simulation scenarios to demonstrate both the variable selection and classification performance of the proposed method. In the first scenario, we simulate data with linear relationships among the views and within a view (see Supplementary Materials). In the second scenario, we simulate the data to show nonlinear relationships. In each scenario, there are $D=2$ views and within each view there are two distinct classes. In all scenarios, we generate 20 Monte Carlo training, tuning, and testing sets. We train the models on the training set, choose optimal  hyper parameters using the tuning set, and obtain classification performance using the testing set. We evaluate the proposed and existing methods using the following criteria: i) test accuracy, and ii) feature selection. For feature selection, we evaluate the methods ability to select the true signals and ignore noise variables. We use true positive rates (TPR), false positive rates (FPR), and F-measure as metrics to evaluate the variable selection performance. In Scenario 1, the first $20$ variables are important, and in Scenario Two, the top $10\%$ of the variables in both views are signals.
% ==================== NONLINEAR SIMULATION ====================

\subsubsection{Nonlinear Simulations}
\begin{figure}[H]
\begin{tabular}{ccccc}
         \centering
         \includegraphics[width=1.0\linewidth]{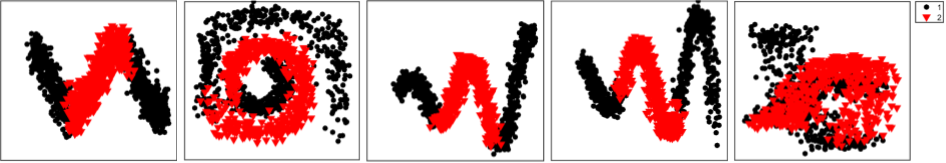}
\end{tabular}
    \caption{Structure of nonlinear relationships between (First left panel) signal variables in View 1; (Second left panel) signal variables in View 2; (Middle panel)-(Fifth panel) signal variables between Views 1 and 2. {Black circle: Class 1; Red triangle: Class 2.}}
    \label{fig:nonlinears1}
\end{figure}

We consider three different settings for this scenario. Each setting has $K=2$ classes, but they vary in dimension. In each setting, $10\%$ of the variables in each view are signals and the first five signal variables in each view are related to the remaining signal variables in a nonlinear way (see \cref{fig:nonlinears1}). We generate data for View 1 as follows: $\mathbf{X}^{(1)}= \widetilde{\bX}_1 \cdot \bW + 0.2\mathbf{E}_1$ where $(\cdot)$ is element-wise multiplication,  $\bW \in \R^{n \times p^{(1)}}= [\mathbf{1}_{0.1\times p^{(1)}}, \mathbf{0}_{0.9\times p^{(1)}}]$ is a matrix of ones and zeros, $\mathbf{1}$ is a matrix of ones, $\mathbf{0}$ is a matrix of zeros, and $\mathbf{E}_1 \sim N(0,1)$. Each of the first five signal variables in $\widetilde{\bX}_1 \in \R^{n \times p^{(1)}}$ is obtained from $\btheta=\tilde{\btheta} + 0.5U(0,1)$, where $\tilde{\btheta}$ is a vector of $n$ evenly spaced points between $0$ and $3\pi$. The next 45 signal variables (or columns) in $\widetilde{\bX}_1 \in \R^{n \times p^{(1)}}$ are simulated from $\cos(\btheta)$ plus random noise from a standard normal distribution.
%Here, $\tilde{\btheta}$ is a vector of $n$ evenly spaced points between $0$ and $3\pi$. 
The remaining $0.9p^{(1)}$ variables (or columns) in $\widetilde{\bX}_1$ are generated from the standard normal distribution. We generate data for View 2 as:
$\mathbf{X}^{(2)}= \widetilde{\bX}_2 \cdot \bW + 0.2\mathbf{E}_2$ where $\mathbf{E}_2 \sim N(0,1)$.  The first five columns (or variables) of $\widetilde{\bX}_2 \in \R^{n \times p^{(2)}}$ are simulated from $\exp(0.15\btheta)\cdot\sin(1.5\btheta)$. The next $0.1p^{(2)} - 5$ variables are simulated from $\exp(0.15\btheta)\cdot\cos(1.5\btheta)$. 
%Here, $\btheta=\tilde{\btheta} + 0.5U(0,1)$, and $\tilde{\btheta}$ is a vector of $n$ evenly spaced points between $0$ and $3\pi$. 
The remaining $0.9p^{(2)}$ variables (or columns) in $\widetilde{\bX}_2$ are generated from the standard normal distribution. The class labels $\mathbf{y} = [\mathbf{1}_{n_1/2}~ 2\cdot\mathbf{1}_{n_2}~ \mathbf{1}_{n_1/2}]$ where $(n_1, n_2)= (200, 150)$ or $(6000, 4500)$. Figure \ref{fig:nonlinears1} shows the structure of the nonlinear relationships between signal variables in View 1(First left panel), signal variables in View 2( Second left panel), and signal varibles between Views 1 and View 2 (Middle to Last panel), with black circles denoting data from Class 1 and red triangles data from Class 2. 

% \begin{figure}[H]
% \begin{tabular}{ccccc}
%          \centering
%          \includegraphics[width=0.18\textwidth]{Plots/deepIDAX1S2.png}& \includegraphics[width=0.18\textwidth]{Plots/deepIDAX2S2.png}& \includegraphics[width=0.18\textwidth]{Plots/deepIDAX1X2_15S2.png}& \includegraphics[width=0.18\textwidth]{Plots/deepIDAX1X2_18S2.png}& \includegraphics[width=0.18\textwidth]{Plots/deepIDAX1X2_10_5S2.png}\\
% \end{tabular}
%     \caption{Structure of nonlinear relationships between (First left panel) signal variables in View 1; (Second left panel) signal variables in View 2; (Middle panel)-(Fifth panel) signal variables between Views 1 and 2.}
%     \label{fig:nonlinears1}
% \end{figure}

\subsubsection{Competing Methods and Results}
We compare the proposed method, iDeepViewLearn, with linear and nonlinear methods for associating data from multiple views. For linear methods, we consider the sparse canonical correlation analysis [Sparse CCA] proposed in \cite{safo2018sparse}. For the nonlinear methods, we compare with deep canonical correlation analysis (Deep CCA) \citep{Andrew:2013}
and MOMA \citep{MOMA:2022}. %, and randomized kernel CCA (RKCCA) \citep{lopez2014randomized}. 
{We note that MOMA is a joint integration and classification method and as such does not require further training a classification method such as SVM, after training MOMA. However, per reviewer comment, we add a comparison where  we use the important features chosen by MOMA to train and test an SVM classifier; we call this MOMA + SVM}. For Sparse CCA and Deep CCA, we use the estimated canonical variates in SVM for  classification performance since these two methods are unsupervised. {We also compare the proposed method that integrates the two views with our method on stacked data, and SVM and random forest~\cite{breiman_random_2001} on stacked data as well, to explore the benefits of multiview learning. Of note, by stacking the data, we do not appropriately model the dependency structure among views as one assumes that the views are not correlated, contrary to the assumption for data integration.} We perform Sparse CCA with the \textit{SelpCCA} R package provided by the authors on GitHub. 
%We perform 
%RKCCA using the Matlab codes provided in the Matlab package for Deep Canonically Correlated Autoencoders\cite{wang2015deep}. We set the number of random features in RKCCA as 300 and we select the bandwidth of the radial basis kernel  using median heuristic.
%\citep{garreau2017large}.  
We performed Deep CCA and MOMA using PyTorch codes provided by the authors. We pair Deep CCA with the teacher-student framework (TS) \citep{TS:2019} to rank variables, and compare the TS feature selection approach with the proposed method. We follow the variable-ranking approach in MOMA to rank variables. 
%\subsubsection{Results}
We report the classification and variable selection results in \cref{tab:nonLinear} for nonlinear simulations (see results of linear settings and the network structures in Supplementary Materials.) We implemented the proposed method in the training data, selected the top $10\%$ variables for each view, learned a new model with these selected variables, and obtained test errors with the test data. The misclassification rates for the proposed method were lower or competitive compared to all the competitors. We observed a decreasing misclassification rate with increasing sample sizes for all the methods; nevertheless, \textbf{the proposed method produced lower or competitive test errors even when the sample size was smaller than the dimension of the variables.}
%From Table \ref{tab:nonLinear}, stacking the views and ignoring the dependency structure among the views resulted in poor classification performance. 
{In terms of variable selection, the TS framework applied to Deep CCA yielded suboptimal results; MOMA and random forest rank the important features based on their influence on the classification performance, and the two methods usually select unimportant features when the sample size is small; \textbf{iDeepViewLearn and Sparse CCA can always achieve nearly perfect performance for feature selection in the nonlinear simulations.} The performance of iDeepViewLearn on the stacked data was similar, although it had slightly higher  classification errors, when compared to iDeepViewLearn that holistically integrates the views; thus we recommended against stacking data and implementing the proposed method, but rather using the method that integrates the two views as we have proposed.} The results of the linear simulations mimic those of the nonlinear simulations.
% \begin{sidewaystable}
\begin{table}

\caption{Nonlinear settings: {randomly select combinations of hyper-parameters to search over}. TPR-1; true positive rate for {$\mathbf{X}^{(1)}$}. Similar for TPR-2. FPR-1; false positive rate for {$\mathbf{X}^{(1)}$}. Similar for FPR-2; F-1 is the F measure for {$\bX^{(1)}$}. Similar for F-2. The highest F-1/2 is in \textcolor{red}{red}. (The mean error of two views is reported for MOMA; {MOMA + SVM means selecting features using MOMA and training an SVM on the selected features.})}

\label{tab:nonLinear}
% (We report the average error from two views for MOMA.)
\begin{center}
\resizebox{\textwidth}{!}{
\begin{tabular}{lrrrrrrr}
\hline
Method&Error (\%)&	TPR-1&TPR-2 & FPR-1 &FPR-2 & F-1& F-2	\\
			\hline
			\hline
			\textbf{Setting 1}&  \\
			$(p_1=500,p_2=500, n_1=200, n_2=150)$\\
	           % \rb{iDeepViewLearn}	&4.77 (1.42)& 100.00	&100.00& 0.00&	0.00& 100.00& 100.00&
	            \textbf{iDeepViewLearn}	&1.89 (0.47)& 100.00	&100.00& 0.00&	0.00& \textcolor{red}{100.00}& \textcolor{red}{100.00} \\
                {iDeepViewLearn on stacked data}	&4.00 (0.47)& 100.00	&100.00& 0.00&	0.00& \textcolor{red}{100.00}& \textcolor{red}{100.00} \\
			    Sparse CCA + SVM & 6.10 (0.73) & 100.00 & 90.00 & 0.11 & 0.01 & 99.51 & 94.69 \\
			    Deep CCA + TS + SVM & 35.61 (2.22)  & 11.10 & 11.30 & 9.88 & 9.86 & 11.10 & 11.30\\
			    %RKCCA&41.21 (4.17)& 	-&-& -&	-& -& -\\
			    % MOMA & 44.78 (1.89)  & 18.50 & 25.90 & 9.06 & 8.23 & 18.50 & 25.90\\
                MOMA & 44.96 (1.70)  & 22.00 & 29.90 & 8.67 & 7.89 & 22.00 & 29.90\\
                {MOMA + SVM} & 30.47 (6.05)  & 22.00 & 29.90 & 8.67 & 7.89 & 22.00 & 29.90\\
                {Random Forest on stacked data} & 1.94 (0.60)  & 70.10 & 98.00 & 3.32 & 0.22 & 70.10 & 98.00\\
			    SVM on stacked data & 28.07 (0.65)  & - & - & - & - & - & -\\
			\hline	
			
            \textbf{Setting 2}&  \\ %originally Setting 3
			$(p_1=500,p_2=500, n_1=6,000, n_2=4,500)$\\
	           % \rb{iDeepViewLearn}	&1.63 (0.15)& 100.00	&100.00& 0.00&	0.00& 100.00& 100.00&
	            \textbf{iDeepViewLearn}	&1.26 (0.11)& 100.00	&100.00& 0.00&	0.00& \textcolor{red}{100.00}& \textcolor{red}{100.00} \\
                {iDeepViewLearn on stacked data}	&1.38 (0.08)& 100.00	&100.00& 0.00&	0.00& \textcolor{red}{100.00}& \textcolor{red}{100.00} \\
			    Sparse CCA + SVM & 4.25 (0.15) & 100.00 & 90.00 & 0.00 & 0.00 & \textcolor{red}{100.00} & 94.74 \\
			    Deep CCA + TS + SVM & 0.66 (0.13) & 30.40 & 21.60 & 7.73 & 8.71 & 30.40 & 21.60\\
			    %RKCCA&11.10 (0.3)& 	-&-& -&	-& -& -\\
                % MOMA & 10.53 (4.77)  & 80.70 & 91.60 & 2.14 & 0.93 & 80.70 & 91.60\\
                MOMA & 12.77 (8.63)  & 76.30 & 89.90 & 2.63 & 1.12 & 76.30 & 89.90\\
                {MOMA + SVM} & 0.63 (0.08)  & 76.30 & 89.90 & 2.63 & 1.12 & 76.30 & 89.90\\
                {Random Forest on stacked data} & 0.66 (0.05) & 100.00 & 100.00 & 0.00 & 0.00 & \textcolor{red}{100.00} & \textcolor{red}{100.00}\\
			    SVM on stacked data & 2.31 (0.15) & - & - & - & - & - & -\\
            \hline
            
            \textbf{Setting 3}&  \\ %originally Setting 4
			$(p_1=2,000,p_2=2,000, n_1=200, n_2=150)$\\
			 %   \rb{iDeepViewLearn}	&5.07 (1.11)& 100.00	&100.00& 0.00&	0.00& 100.00& 100.00&
			    \textbf{iDeepViewLearn}	&2.56 (0.78)& 99.98	&99.88& 0.00&	0.00& \textcolor{red}{99.98}	&\textcolor{red}{99.88} \\
                {iDeepViewLearn on stacked data}	&2.86 (0.73)& 99.98	&99.65& 0.01&	0.04& \textcolor{red}{99.98}	&99.65 \\
			    Sparse CCA + SVM & 4.86 (0.88) & 100.00 & 97.50 & 0.08 & 0.02 & 99.63 & 98.66 \\
			    Deep CCA + TS + SVM & 29.91 (1.27)  & 10.30 & 11.20 & 9.97 & 9.87 & 10.30 & 11.20\\
			    %RKCCA&43.20 (2.64)&	1.13&0& -&	-& -& -\\
                % MOMA & 45.46 (2.40)  & 17.12 & 13.90 & 9.21 & 9.57 & 17.12 & 13.90\\
                MOMA & 46.14 (2.44)  & 16.40 & 13.68 & 9.29 & 9.59 & 16.40 & 13.68\\
                {MOMA + SVM} & 35.46 (5.91)  & 16.40 & 13.68 & 9.29 & 9.59 & 16.40 & 13.68\\
                {Random Forest on stacked data} & 5.40 (1.02) & 58.67 & 89.88 & 4.59 & 1.13 & 58.67 & 89.88\\
			    SVM on stacked data & 28.57 (0.53) & - & - & - & - & - & -\\
            \hline
            \hline
            
\end{tabular}}
\end{center}
\end{table}

% ==================== GRAPH SIMULATION ====================

\subsection{Set-up when there is prior information on variable-variable interactions}
\begin{figure}[!htbp]
    \centering
    \includegraphics[width=0.3\linewidth]{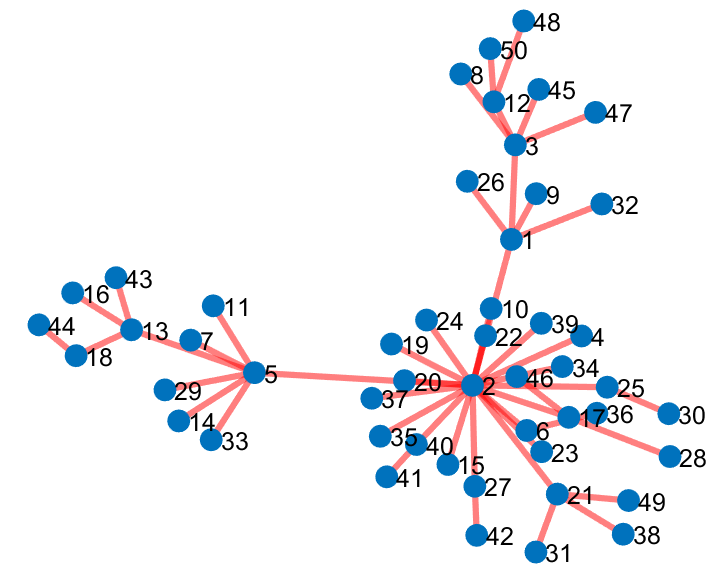} \includegraphics[width=0.3\linewidth]{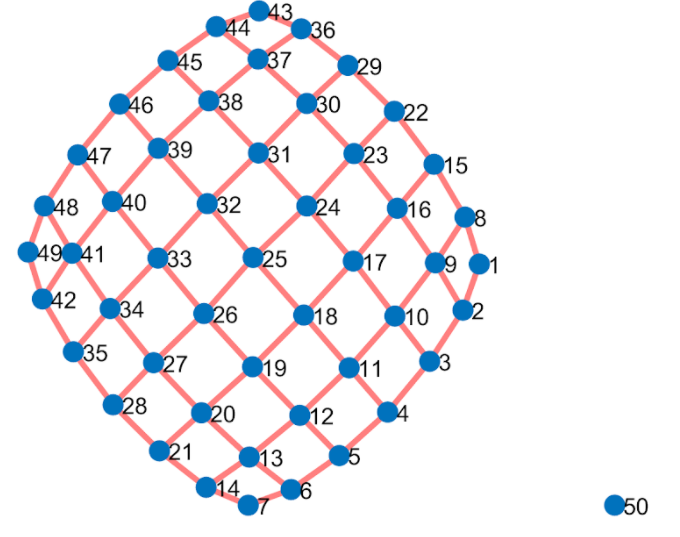} \includegraphics[width=0.3\linewidth]{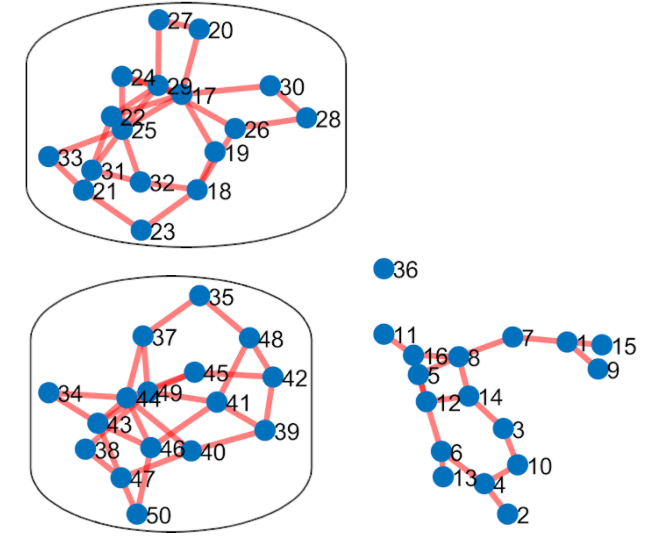}\\
%    (A) ~~~~& (B) ~~~~& C
    \caption{Network structure for the first 50 variables in $\bX^{(1)}$ and $\bX^{(2)}$. Left: scale-free network; Middle: Lattice; Right: Cluster. For the Scale-free network, we consider variable 2 has a hub variable. Variable 2 and the variables directly connected to it are considered as signal variables. For the Lattice network, all variables except variable 50 are considered as signals. For the Cluster network, the circled clusters are considered as signals. }
    \label{fig:network}
\end{figure}

Here,  $\mathbf{X}^{(1)}= \widetilde{\bX}_1 \cdot \bW + \mathbf{E}_1$ and $\mathbf{X}^{(2)}= \widetilde{\bX}_2 \cdot \bW + 0.2\mathbf{E}_2$. $\widetilde{\bX}_i$, $i=1,2$ is defined as before.  However, $\mathbf{E}_i \sim N(0, \bSigma_i$), $i =1,2$, $\bSigma_i$ is a diagonal block matrix with two blocks that represent signal and noise variables. The first block is a $50 \times 50$ covariance matrix that captures the relationship among these 50 variables. Let ${G}$ be the true graph structure for these variables. The second block is the identity matrix.  We use \textit{bdgraph.sim} in the \textit{BDGraph} R package \citep{mohammadi2015bdgraph} to simulate three different types of networks for the first 50 variables: Scale-free, Lattice, and Cluster, and to obtain the adjacency matrix corresponding to the graph structure ${G}$. We then use \textit{rgwish} from the same R package to generate a precision matrix distributed according to the $G-$Wishart distribution, $W_G(b,D)$, with parameters $b=3$ and $D=\bI$ with respect to the graph structure $G$.  We obtain the covariance matrix from the precision matrix. \cref{fig:network} shows the variable-variable relationships among these 50 variables for the different network structure. \cref{fig:network} [left panel], variable two is connected to more variables, so we consider variable 2 as a hub variable. We set $\mathbf{W}= [\mathbf{1}_{\mathcal{H}}, \mathbf{0}_{p^{(1)}-\mathcal{H}}]$, $\mathcal{H}$ to denote the variables directly connected to variable 2, and $p^{(1)}-\mathcal{H}$ (similarly $p^{(2)} - \mathcal{H}$) denote the remaining variables. By defining $\mathbf{W}$ this way, we assume that only the variables directly connected to the hub variable are signals and contribute to the nonlinear relationship between the two views. For the Lattice network (Middle panel), all variables in the network except variable 50 contribute to the nonlinear relationships among the views. For the cluster network, only two clusters (circled) are signals.

\subsubsection{Competing Methods and Results}
We explore the proposed method with and without the use of network information. In addition to competitors in the nonlinear simulations, we further compare the proposed method with Fused CCA \citep{safo2018integrative}. Fused CCA is a sparse canonical correlation analysis method that uses variable-variable information to guide the estimation of the canonical variates and the selection of variables that contribute most to the association between two views. We implemented Fused CCA using the R code accompanying the manuscript. We followed Fused CCA with SVM for classification. We implemented the proposed method on the training data with and without the network information, selected the top ranked variables (21 variables for Scale-free network, 49 variables for Lattice network, and 33 for Cluster network) for each view, learned a new model with these selected variables, and obtained test errors with the testing data. We also compared the top-ranked variables with the true signal variables and the estimated true positive rates (TPR), false positive rates (FPR), and F-score. From \cref{tab: nonLinearGraph}, the classification performance of our method that does not incorporate prior knowledge is comparable to our method that does, in all settings. The fused CCA result for Scale-free Setting 2 was based on 19 out of the 20 simulation replicates due to a computational error. The fused CCA result for Lattice Setting 2 was not available due to time constraints. \textbf{Like the scenario with no prior information, the misclassification rates for the proposed method (with or without prior information) were lower or competitive, especially for the Scale-free and Lattice networks, when compared to the association-based methods.} Furthermore, the proposed method was superior to MOMA and SVM on stacked views across all settings and network type, for both classification and variable selection accuracy. {For random forest, it can achieve very comparable prediction and feature selection performance with our methods when there are sufficient training data points; however, \textbf{our iDeepViewLearn's performance outperforms  random forest when the sample size is limited when considering both classification and variable selection performance. }}

In summary, the classification and variable selection accuracy from both the linear and nonlinear simulations, and when we use or do not use prior information, suggest that the proposed methods are capable of ranking signal variables as high and ignoring noise variables. The proposed methods are also capable of producing competitive or better classification performance among all settings. {In particular, we notice that random forest can achieve comparable classification and variable selection accuracies with our iDeepViewLearn when the number of training samples is relatively large, but the feature selection performance of random forest is usually suboptimal in situations where the sample size is less than the number of variables, as shown in~\cref{tab:nonLinear,tab: nonLinearGraph}. These findings are encouraging to us since in a typical setting of high-dimensional and biomedical problems, the sample size is smaller than the number of variables. }

% \begin{figure}[!htbp]
%     \centering
%     \includegraphics[width=0.3\linewidth]{Plots/scalefreenetwork2.png} \includegraphics[width=0.3\linewidth]{Plots/lattice2.png} \includegraphics[width=0.3\linewidth]{Plots/cluster3.png}\\
% %    (A) ~~~~& (B) ~~~~& C
%     \caption{Network structure for the first 50 variables in $\bX^{(1)}$ and $\bX^{(2)}$. Left: scale-free network; Middle: Lattice; Right: Cluster. For the Scale-free network, we consider variable 2 has a hub variable. Variable 2 and the variables directly connected to it are considered as signal variables. For the Lattice network, all variables except variable 50 are considered as signals. For the Cluster network, the circled clusters are considered as signals. }
%     \label{fig:network}
% \end{figure}

\begin{table}

\caption{Simulation with variable-variable connections: {randomly select combinations of hyper-parameters to search over}. TPR-1; true positive rate for {$\mathbf{X}^{(1)}$}. Similar for TPR-2. FPR; false positive rate for {$\mathbf{X}^{(2)}$}. Similar for FPR-2; F-1 is the F measure for {$\bX^{(1)}$}. Similar for F-2. The highest F-1/2 is in \textcolor{red}{red}. (The mean error of two views is reported for MOMA; {MOMA + SVM means combining the feature selection part of MOMA and SVM.})}
\vspace{-0.5em}
\label{tab: nonLinearGraph}
\begin{center}
\resizebox{\textwidth}{!}{
\begin{tabular}{lrrrrrrr}
\hline
Method&Error (\%)&	TPR-1&TPR-2 & FPR-1 &FPR-2 & F-1& F-2	\\
			\hline
			\hline
			\textbf{Scale-free}&   \\
			    \textbf{Setting 1} &  \\$(p_1=500,p_2=500, n_1=200, n_2=150)$
                \\
	           %     \textbf{iDeepViewLearn}	&10.71 (2.61)& 94.29	&98.33& 0.25&	0.07& 90.99& 93.04 \\
		          %  \textbf{iDeepViewLearn-Laplacian}	&10.44 (1.88)& 97.14	&99.76& 0.13&	0.01& 91.12& 92.71 \\
		            
		            \textbf{iDeepViewLearn}	&6.84 (1.96)& 95.71	&96.19& 0.19&	0.17& {95.71}	&{96.19} \\
		            % \textbf{iDeepViewLearn-Laplacian}	&11.77 (2.52)& 73.57	&89.52& 1.16&	0.46& 73.57	&89.52 \\

                    \textbf{iDeepViewLearn-Laplacian}	&7.10 (1.65)& 97.86	&98.57& 0.09&	0.06& \textcolor{red}{97.86}	&\textcolor{red}{98.57} \\
		            %\textcolor{orange}{SIDANet} & 47.73 (2.07) & 12.14 & 10.71 & 9.70 & 9.28 & 6.06 & 4.69 \\
				    Sparse CCA + SVM & 21.80 (10.14)  & 100.00 & 100.00 & 14.12 & 15.13 & 42.97 & 43.26 \\
                    Fused CCA + SVM & 41.33 (7.52) & 19.29 & 23.10 & 17.90 & 33.29 & 6.56 & 4.08 \\
			        Deep CCA + TS + SVM & 40.43 (1.61)  & 5.00 & 3.33 & 4.16 & 4.24 & 5.00 & 3.33\\
                    MOMA & 45.55 (1.84)  & 17.14 & 25.95 & 3.63 & 3.25 & 17.14 & 25.95\\
                    {MOMA + SVM} & 36.76 (5.41)  & 17.14 & 25.95 & 3.63 & 3.25 & 17.14 & 25.95\\
                    {Random Forest on stacked data} & 11.99 (2.22)  & 52.38 & 85.23 & 2.09 & 0.65 & 52.38 & 85.23\\
			        SVM on stacked data & 35.46 (1.23)  & - & - & - & - & - & -\\
			    \hline				

                \textbf{Setting 2}&  \\
                $(p_1=500,p_2=500, n_1=6,000, n_2=4,500)$\\
	               % \textbf{iDeepViewLearn}	&3.73 (0.34)& 100.00	&99.76& 0.00&	0.01& 96.73& 96.69&
	               % \textbf{iDeepViewLearn-Laplacian}	&3.78 (0.52)& 100.00	&100.00& 0.00&	0.00& 96.69& 96.70&
	                \textbf{iDeepViewLearn}	&2.77 (0.62)& 99.76	&100.00& 0.01&	0.00& {99.76}& \textcolor{red}{100.00} \\
		            % \textbf{iDeepViewLearn-Laplacian}	&3.45 (1.52)& 95.48	&100.00& 0.20&	0.00& 95.48& \textcolor{red}{100.00} \\

                    \textbf{iDeepViewLearn-Laplacian}	&2.71 (0.14)& 100.00	&100.00& 0.00&	0.00& \textcolor{red}{100.00}	&\textcolor{red}{100.00} \\
                    %\textcolor{orange}{SIDANet} &53.22  (0.83) & 99.29 & 86.67 & 3.52 & 2.57 & 71.04 & 70.59 \\
			        Sparse CCA + SVM & 9.25 (2.39)&100.00 & 100.00	&9.35& 9.22&	56.74& 52.59 \\
                    Fused CCA + SVM & 33.24 (4.83) & 99.75 & 100.00 & 13.69 &  48.83 & 48.74 & 19.90 \\
			        Deep CCA + TS + SVM & 2.44 (0.31)  & 17.62 & 17.38 & 3.61 & 3.62 & 17.62 & 17.38\\
                    MOMA & 41.88 (4.11)  & 63.81 & 72.62 & 1.59 & 1.20 & 63.81 & 72.62\\
                    {MOMA + SVM} & 3.56 (4.37)  & 63.81 & 72.62 & 1.59 & 1.20 & 63.81 & 72.62\\
                    {Random Forest on stacked data} & 1.86 (0.10)  & 100.00 & 100.00 & 0.00 & 0.00 & \textcolor{red}{100.00} & \textcolor{red}{100.00}\\
			        SVM on stacked data & 27.61 (0.23)  & - & - & - & - & - & -\\
                \hline
            \hline
			\textbf{Lattice}&   \\			
			    \textbf{Setting 1}&  \\
			    $(p_1=500,p_2=500, n_1=200, n_2=150)$\\
	           %     \textbf{iDeepViewLearn}	&4.94 (1.83)& 98.88	&97.35& 0.12&	0.29& 95.69& 95.90&
		          %  \textbf{iDeepViewLearn-Laplacian}	&6.31 (3.14)& 97.45	&95.71& 0.28&	0.47& 94.51& 94.86&
		            \textbf{iDeepViewLearn}	&4.90 (1.76)& 100.00	&98.88& 0.00&	0.12& \textcolor{red}{100.00}& 98.88 \\
		            % \textbf{iDeepViewLearn-Laplacian}	&4.87 (0.99)& 99.18	&100.00& 0.09&	0.00& 99.18& \textcolor{red}{100.00} \\

                    \textbf{iDeepViewLearn-Laplacian}	&3.90 (0.82)& 99.80	&99.59& 0.02&	0.04& 99.80	&\textcolor{red}{99.59} \\
                    %\textcolor{orange}{SIDANet} & 47.09 (2.80) & 11.43 & 4.69 &10.25 & 8.39  & 8.18 & 4.02 \\
			        Sparse CCA + SVM & 16.03 (0.86)& 100.00 & 100.00 & 1.29 & 1.45 & 94.96 & 94.77   \\
                    Fused CCA + SVM & 38.26 (11.58) & 22.04 & 24.69 & 24.59 & 30.99 & 11.13 & 9.30 \\
			        Deep CCA + TS + SVM & 36.53 (2.05) & 10.61 & 10.71 & 9.71 & 9.70 & 10.61 & 10.71\\
                    MOMA & 44.76 (2.12)  & 23.98 & 26.53 & 8.26 & 7.98 & 23.98 & 26.53\\
                    {MOMA + SVM} & 32.20 (4.27)  & 23.98 & 26.53 & 8.26 & 7.98 & 23.98 & 26.53\\
                    {Random Forest on stacked data} & 3.41 (0.71) & 66.84 & 92.86 & 3.60 & 0.78 &  66.84 & 92.86 \\
			        SVM on stacked data & 28.51 (0.56) & - & - & - & - & - & - \\
			    \hline				

                \textbf{Setting 2}&  \\
				$(p_1=500,p_2=500, n_1=6,000, n_2=4,500)$\\
	               % \textbf{iDeepViewLearn}	&2.85 (0.24)& 100.00	&100.00& 0.00&	0.00& 97.51& 97.54 \\
	               % \textbf{iDeepViewLearn-Laplacian}	&2.83 (0.25)& 100.00	&100.00& 0.00&	0.00& 97.52& 97.55 \\
	                \textbf{iDeepViewLearn}	&1.64 (0.17)& 100.00	&100.00& 0.00&	0.00& \textcolor{red}{100.00}& \textcolor{red}{100.00} \\
		            % \textbf{iDeepViewLearn-Laplacian}	&1.95 (0.15)& 100.00	&100.00& 0.00&	0.00& \textcolor{red}{100.00}& \textcolor{red}{100.00} \\

                    \textbf{iDeepViewLearn-Laplacian}	&1.56 (0.12)& 100.00	&100.00& 0.00&	0.00& \textcolor{red}{100.00}& \textcolor{red}{100.00} \\
                    %\textcolor{orange}{SIDANet} &48.87 (0.49) & 35.00 & 95.71 & 0.01 & 14.40 & 47.77 & 60.76 \\
			        Sparse CCA + SVM & 7.14 (2.92)&100.00 & 100.00	&1.14& 1.72&	95.52& 93.52 \\
                    Fused CCA + SVM & 5.26 (2.27) & 100.00 & 100.00 & 3.44 & 6.84 & 88.89 & 78.95 \\
			        Deep CCA + TS + SVM & 0.98 (0.19)  & 39.49 & 32.04 & 6.57 & 7.38 & 39.49 & 32.04\\
                    MOMA & 21.22 (13.01)  & 73.98 & 79.08 & 2.83 & 2.27 & 73.98 & 79.08\\
                    {MOMA + SVM} & 1.27 (1.53)  & 73.98 & 79.08 & 2.83 & 2.27 & 73.98 & 79.08\\
                    {Random Forest on stacked data} & 1.02 (0.07) & 100.00 & 100.00 & 0.00 & 0.00 &  \textcolor{red}{100.00} & \textcolor{red}{100.00}\\
			        SVM on stacked data & 8.57 (0.24) & - & - & - & - & - & -\\
                \hline
            \hline
			\textbf{Cluster}&   \\			
			    \textbf{Setting 1}&  \\
			    $(p_1=500,p_2=500, n_1=200, n_2=150)$\\
	           %     \textbf{iDeepViewLearn}	&25.56 (0.49)& 85.00	&90.61& 1.06&	0.66& 79.44& 83.10 \\
		          %  \textbf{iDeepViewLearn-Laplacian}	&24.00 (4.01)& 87.12	&92.27& 0.91&	0.55& 80.38& 83.58 \\
		            \textbf{iDeepViewLearn}	&22.50 (1.73)& 96.21	&100.00& 0.27&	0.00& \textcolor{red}{96.21}& \textcolor{red}{100.00} \\
		            % \textbf{iDeepViewLearn-Laplacian}	&22.91 (1.85)& 93.03	&100.00& 0.50&	0.00& 93.03& \textcolor{red}{100.00} \\

                    \textbf{iDeepViewLearn-Laplacian}	&22.40 (2.14)& 95.15	&100.00& 0.34&	0.00& 95.15	&\textcolor{red}{100.00} \\
                    %\textcolor{orange}{SIDANet} & 47.54 (3.40) & 12.88 & 5.45 & 10.12 & 8.77 & 7.37 & 3.74 \\
				    Sparse CCA + SVM & 16.70 (1.22)  & 100.00 & 100.00 & 4.24 & 3.91 & 77.50 & 78.54 \\
                    Fused CCA + SVM & 43.27 (1.65) & 16.97 & 16.06 & 18.96 & 22.76 & 5.52 & 5.56 \\
			        Deep CCA + TS + SVM & 37.96 (1.81) & 7.12 & 6.97 & 6.56 & 6.57 & 7.12 & 6.97 \\
                    MOMA & 45.28 (2.02)  & 21.52 & 23.18 & 5.55 & 5.43 & 21.52 & 23.18\\
                    {MOMA + SVM} & 36.61 (3.77)  & 21.52 & 23.18 & 5.55 & 5.43 & 21.52 & 23.18\\
                    {Random Forest on stacked data} & 29.23 (1.19)  & 27.42 & 65.76 & 5.13 & 2.42 & 27.42 & 65.76\\
			        SVM on stacked data & 31.60 (1.02)  & - & - & - & - & - & -\\
			    \hline				
                
                \textbf{Setting 2}&  \\
				$(p_1=500,p_2=500, n_1=6,000, n_2=4,500)$\\
	               % \textbf{iDeepViewLearn}	&19.14 (1.51)& 96.52	&100.00& 0.25&	0.00& 83.79& 85.40&
	               % \textbf{iDeepViewLearn-Laplacian}	&20.58 (2.56)& 88.63	&94.09& 0.80&	0.42& 82.51& 83.88&
	                \textbf{iDeepViewLearn}	&15.78 (0.65)& 100.00	&99.39& 0.00&	0.04& \textcolor{red}{100.00}	&{99.39} \\
		            % \textbf{iDeepViewLearn-Laplacian}	&16.63 (0.82)& 90.61	&96.06& 0.66&	0.28& 90.61	&96.06 \\

                    \textbf{iDeepViewLearn-Laplacian}	&15.70 (0.35)& 96.21	&100.00& 0.27&	0.00& 96.21	&\textcolor{red}{100.00} \\
                    %\textcolor{orange}{SIDANet} & 50.40 (1.47) & 98.64 & 74.09 & 0.63 & 0.00 & 95.10 & 84.84 \\
		          Sparse CCA + SVM & 14.59 (0.53) &100.00 & 100.00	&12.07& 7.54&	57.31& 66.46 \\
                    Fused CCA + SVM & 29.17 (9.55) & 72.73 & 92.42 & 26.12 & 30.40 & 31.17 & 41.73 \\
			        Deep CCA + TS + SVM & 28.48 (1.52)  & 10.45 & 8.64 & 6.33 & 6.46 & 10.45 & 8.64\\
                    MOMA & 39.22 (4.95)  & 73.18 & 91.82 & 1.90 & 0.58 & 73.18 & 91.82\\
                    {MOMA + SVM} & 12.77 (0.72)  & 73.18 & 91.82 & 1.90 & 0.58 & 73.18 & 91.82\\
                    {Random Forest on stacked data} & 13.83 (0.21)  & 100.00 & 100.00 & 0.00 & 0.00 & \textcolor{red}{100.00} & \textcolor{red}{100.00}\\
			        SVM on stacked data & 29.68 (0.22)  & - & - & - & - & - & -\\
\hline
\hline
\end{tabular}}
\end{center}
\end{table}

% ==================== REAL DATA ====================
% \section{Analysis of Data from the Holm Breast Cancer Study}
\section{{Real-World Experiments}}
\label{s: real}

\begin{table}[]
    \centering
    \caption{{Summary of datasets for each analysis.}}
    \resizebox{\textwidth}{!}{
    \begin{tabular}{lllll}
    \hline
    \hline
        Dataset & Categories & Number of features  & Sample Size & Task \\
                &            & in each view        &  & \\
        \hline
        Holm Breast Cancer & Died: 65 & View 1, gene expression, 469 & Training $n=112$ & Classification and  \\
        Study              & Survived: 103 & View 2, methylation, 334     & Testing $n=56$  & Clustering\\
        \hline
        LGG Dataset & Grade 2: 246 & View 1, methylation, 9691 & Training $n=410$ & Classification \\
                    & Grade 3: 264 & View 2, miRNA, 235        & Testing $n=100$  & \\
                    &              & View 3, mRNAseq, 7603 & & \\
        \hline
        Shear Transformed & Hand-written digits 0 to 9 & View 1, digits, 784 & Training $n=60000$ & Classification and  \\
        MNIST Dataset & count ranging from 5400 to 6800 & View 2, digits, 784 & Testing $n=10000$ & Reconstruction\\
        
    \hline
    \hline
    \end{tabular}
    }
    \label{tab:dataset_info}
\end{table}
{In this section, we consider three real-world applications to show the effectiveness of the proposed method across different tasks and settings. In \cref{s: real_bc}, we applied the proposed method to integrate gene expression and methylation data from the Holm breast cancer study \citep{holm2010molecular} for classification and clustering tasks with two views. In \cref{s: real_lgg}, we applied the proposed method to data pertaining to brain lower grade glioma (LGG) to demonstrate the use of our method for classification tasks with three views. In \cref{s: real_mnist}, we used MNIST handwriting data, for a reconstruction task, demonstrating that handwriting digits can be reconstructed with few pixels while maintaining competitive classification accuracy. The details of all the datasets used in this section are shown in \cref{tab:dataset_info}.}

\subsection{{Evaluation of Data from Holm Breast Cancer Study}}
\label{s: real_bc}
Breast cancer is the most common cancer among women worldwide, accounting for 12.5\% of new cases and is one of the leading causes of death in women \citep{giaquinto2022breast}. Research shows that breast cancer is a multi-step process that involves both genetic and epigenetic changes. Epigenetic factors such as DNA methylation and histone modification lead to breast tumorigenesis by silencing critical tumor suppressor and growth regulator genes \citep{lustberg2011epigenetic}. Identifying methylated sites correlated with gene expression data could shed light on the genomic architecture of breast cancer. Our work is motivated by a molecular subtyping study conducted in \cite{holm2010molecular}, which used gene expression and DNA methylation data to identify methylation patterns in breast cancer. For completeness, we describe the data here. Raw methylation profiles from 189 breast cancer samples were extracted using the
Beadstudio Methylation Module (Illumina). There were 1,452 CpG sites (corresponding to 803 cancer-related genes). $\beta$-values were stratified into three groups: 0, 0.5, and 1. The value of 1 corresponded to hypermethylation. Relative methylation levels were obtained from raw values by centering the stratified $\beta$ values in all samples. Furthermore, relative gene expression levels of $179$ of $189$ breast cancer tumors were obtained using oligonucleotide arrays for 511 probes. The number of samples with data on gene expression and methylation for our analysis is $n=179$. The first view, corresponding to gene expression data, had $468$ variables (genes), and the second view, corresponding to methylation data, had $1,452$ variables (CpG sites). The methylation data were filtered to include the most variable methylated sites by removing CpG sites with a standard deviation less than $0.3$ between samples; this resulted in 334 CpG sites (corresponding to 249 cancer-related genes). 
%Following \cite{Holm}, we did not preprocess the gene expression data. 
%The samples have been classified into molecular subtypes as described in \cite{Holm2010molecular}. 
In addition to molecular data, data on whether an individual died from breast cancer or not is also available. 
%Our analytical data were: $\mathbf{X}^{(1)}_{168 \times 468}$ and $\mathbf{X}^{(2)}_{168 \times 334}$.

The goal of our analysis is to perform an integrative analysis of the methylation and gene expression data to model nonlinear associations between CpG sites and genes through a joint non-linear embedding that drives the overall dependency structure in the data. Importantly, we wish to identify a subset of CpG sites and genes that contribute to the dependency structure and could be used to discriminate between those who survived and those who did not survive breast cancer. Further, we wish to explore the use of the joint nonlinear embedding in molecular clustering. 
\subsubsection{Goal 1: Model nonlinear relationships between methylation and gene expression data and identify CpG sites and genes that can discriminate between those who died and those who did not die from breast cancer} 

%\noindent{\textbf{\textit{Application of the proposed and competing methods}}}
For the first goal, we split the data into three sets of approximately equal size. We used $2/3$rd of the data to train the model and we used the remaining $1/3$rd to test our models. We implemented the proposed method in the training set, selected the top $10\%$ and $20\%$ highly ranked variables in each view, learned new models with these selected features, used the test data and the models learned to predict the test outcomes, and obtained test errors. We repeated the process 20 times, obtained the highly ranked variables for each run, and estimated the average test errors. We compared the proposed method with SVM, random forest, Deep CCA, Sparse CCA, MOMA and MOMA + SVM based on average test errors. 

\noindent{\textbf{\textit{Average misclassification rates and genes and CpG sites selected}}}
\cref{tab: breastcancer} gives the average test errors for the methods. On the basis of the high classification errors across the methods, it seems that separating those who died from breast cancer from those who did not die using methylation and gene expression data is a difficult problem. {We investigate the use of iDeepViewLearn on the stacked data, and we notice that, similar to the nonlinear simulations, there are small gaps present compared to the results when we integrate the data. Hence, we still recommend using our method as proposed when there are two or more views, and not used on stacked data.} The average test error for the proposed method based on the top $10\%$ or $20\%$ CpG sites and genes is comparable to that of the other methods. Our proposed method allows us to obtain insight into the genes and CpG sites that drive the classification accuracy.

\begin{table}
\caption{Breast cancer data: SVM is based on stacked raw data with two views. Deep CCA + SVM is a training SVM based on the last layer of Deep CCA. iDeepViewLearn with selected top $10\%$ features reconstructs the original views with only $10\%$ of the features and obtains a test classification error based on a shared low-dimensional representation trained on data with only $10\%$ of the features. Similar to iDeepViewLearn with selected top $20\%$. (The mean error of two views is reported for MOMA; {MOMA + SVM means combining the feature selection part of MOMA and SVM.})}
\label{tab: breastcancer}
\begin{center}
\begin{tabular}{lr}
\hline
Method&Average Error (Std.Dev) (\%)  	\\
			\hline
			\hline
            SVM & 39.02 (4.77)~\\
            %SVM on top 10\% features selected by Deep CCA& 41.43 (5.22)\\
            %SVM on top 10\% features selected by Deep Omics & 41.07 (3.87)\\
			Deep CCA + SVM & 38.57 (5.40)\\
%			Deep CCA + SVM on top 10\% features selected by Deep CCA & 39.91 (4.93)\\
%			Deep CCA + SVM on top 10\% features selected by Deep Omics & 40.71 (4.61)\\
			Sparse CCA + SVM & 40.94 (4.24) \\
            MOMA & 44.51 (3.90) \\
            {MOMA + SVM} & 39.46 (5.67) \\
            {Random Forest} & 40.36 (5.28) \\
%			SIDA & 43.84 (1.28)\\
%			SIDA (two views)&  \\
			\textbf{iDeepViewLearn} with selected top 10\% features & 39.02 (5.03)\\
			\textbf{iDeepViewLearn} with selected top 20\% features & 39.02 (5.03)\\
           {iDeepViewLearn with selected top 10\% stacked features} & 39.11 (4.82)\\
           {iDeepViewLearn with selected top 20\% stacked features} & 39.38 (5.55)\\
\hline
\end{tabular}
\end{center}
\end{table}

For this purpose, we explored the ``stable"  genes and CpG sites that potentially discriminate between people who died and those who did not die from breast cancer. We consider a variable to be ``stable" if it is ranked in the top 20\% at least 16 times ($\ge 80\%$) of the 20 resampled datasets.  
%For robustness in variables selected and to reduce false findings in identifying highly-ranked genes and CpG sites that potentially discriminate persons who died from those who did not die from breast cancer, we used resampling techniques to identify variables that were  ranked in the top 20\% at least 16 times ( $\ge 80\%$) out of the 20 resampled datasets.  
\cref{tab: breastcancerVarSelGenes} shows the genes selected and how often they were selected. Genes DAB2, DCN, HLAF, MFAP4, MMP2, PDGFRB, TCF4 and TMEFF1 were consistently selected in the top $20\%$ in all resampled data sets. There is support in the literature for the potential role of some of these genes in a variety of human cancers. Disabled homolog 2 [or DAB adaptor protein 2] (DAB2) is a protein-coding gene that is often deleted or silenced in several human cancer cells.  
%We found that patients who died from breast cancer had XXX levels of this gene compared to patients who did not die from breast cancer (Figure XXX). 
The decorin gene (DCN) is a protein coding gene that encodes the protein decorin. Research on different human cancers (e.g. breast, prostate, bladder) has shown that DCN expression levels in cancerous cells are significantly reduced from expression levels in normal tissues or are often completely silenced in tumor tissues \citep{jarvinen2015decorin}. Research suggests that individuals expressing lower levels of DCN in cancer tend to have poorer outcomes compared to individuals expressing higher levels of DCN. In our data, the mean expression levels of DCN for those who survived were not statistically different (based on the Anova test) from those who did not.

\begin{table}
% \spacingset{1}

\centering
%	\begin{scriptsize}
				\caption{Frequency of Genes selected at least 16 times in the top 20\% across 20 resampled datasets \label{tab: breastcancerVarSelGenes}}
    \resizebox{\textwidth}{!}{
				\begin{tabular}{lll}
			\hline
			\hline
			Gene&  Gene Name & Frequency	\\
			\hline
			\hline
%			Gene Exp	Gene Name	Frequency
DAB2&	DAB adaptor protein 2&	20\\
DCN	&decorin&	20\\
HLAF&	major histocompatibility complex, class I, F&	20\\
MFAP4&	microfibril associated protein 4&	20\\
MMP2&	matrix metallopeptidase 2&	20\\
PDGFRB&	platelet derived growth factor receptor beta&	20\\
TCF4&	transcription factor 4&	20\\
TMEFF1&	transmembrane protein with EGF like and two follistatin like domains 1&	20\\
AFF3&	AF4/FMR2 family member 3&	19\\
BIRC5&	baculoviral IAP repeat containing 5	&19\\
CDH11&	cadherin 11	&19\\
COL1A2&	collagen type I alpha 2 chain&	19\\
LYN	&LYN proto-oncogene, Src family tyrosine kinase&	19\\
SPARC&	secreted protein acidic and cysteine rich&	19\\
THBS2&	thrombospondin 2&	19\\
BGN	&biglycan&	18\\
COL6A1&	collagen type VI alpha 1 chain&	18\\
CSPG2&	versican&	18\\
LOX	&lysyl oxidase	&18\\
SLIT2&	slit guidance ligand 2&	18\\
TIMP2&	TIMP metallopeptidase inhibitor 2&	18\\
EPHB3&	EPH receptor B3	&17\\
HLADPA1&	Major Histocompatibility Complex, Class II, DP Alpha 1&	17\\
IGFBP7	&insulin like growth factor binding protein 7&	17\\
SPDEF	&SAM pointed domain containing ETS transcription factor&	17\\
THY1&	Thy-1 cell surface antigen&	17\\
TNFRSF1B&	TNF receptor superfamily member 1B&	17\\
IL16	&interleukin 16&	16\\
\hline
			\hline
		\end{tabular}}
%	\end{scriptsize}	
\end{table}	

\begin{figure}[!htbp]
    \centering
    \includegraphics[width=1.0\linewidth]{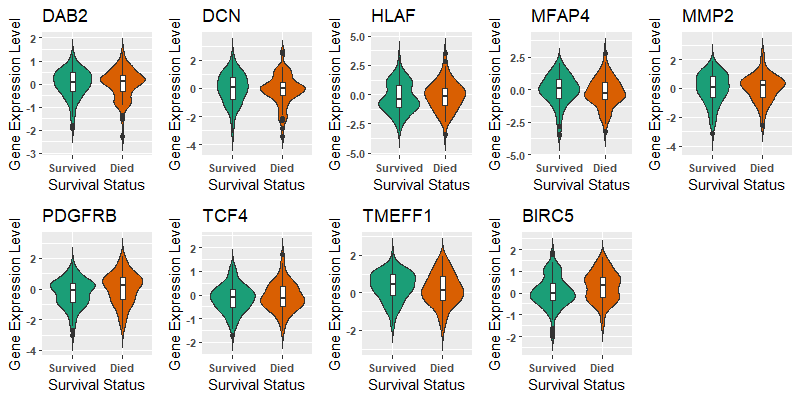}
    \caption{All genes except BIRC5 were consistently selected in the top $20\%$ of highly-ranked genes across the twenty resampled datasets. BIRC5 was selected 19 times (out of 20) in the top $20\%$ highly-ranked genes. Genes PDGFRB and BIRC5 have mean expression levels that are statistically significantly different between individuals that died from breast cancer and those that survived. }
    \label{fig:consistentAnovaSig}
\end{figure}

We observed statistically significant differences in mean expression levels of PDGFR and BIRC5 for the two groups (p-value$< 0.05$ from the ANOVA test), as shown in \cref{fig:consistentAnovaSig}. The mean (median) expression values of these genes were higher in individuals who died of breast cancer. The platelet-derived growth factor receptor alpha (PDGFRA) gene is a protein encoding gene that encodes the PDGFRA protein. The PDGFRA protein is involved in important biological processes such as cell growth, division, and survival.  Mutated forms of the PDGFRA gene and protein have been found in some types of cancer. The BIRC5 gene is a protein encoder gene that encodes the baculoviral IAP repeat containing protein 5 in humans. This protein is believed to play an important role in the promotion of cell division (proliferation) and in the prevention of cell apoptosis (death) \citep{oparina2021prognostic,li1998control}. 

%(cite PubMed:9859993, PubMed:21364656, PubMed:20627126, PubMed:25778398, PubMed:28218735). 
%Further, accumulation of research points to the potential use of DCN as anticancer therapeutic target \cite{jarvinen2015decorin}. 

\begin{figure}[!htbp]
    \centering
    \includegraphics[width=0.9\linewidth]{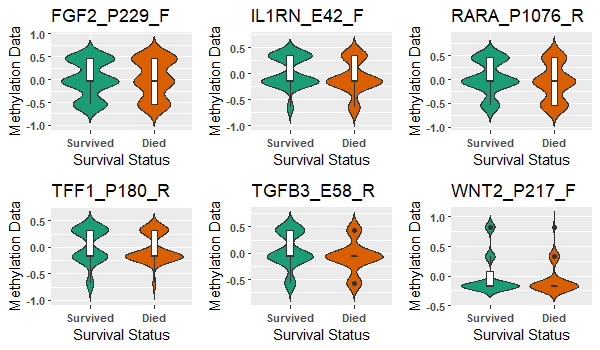}
    \caption{All CpG sites were consistently selected in the top 20\% of highly-ranked CpG sites across the twenty resampled datasets. The mean methylation levels of IL1RN\_E42\_F and TGFB3\_E58\_R  are  statistically  different between individuals that died from breast cancer and those that survived. }
    \label{fig:consistentMethAnovaSig}
\end{figure}

\cref{tab: breastcancerVarSelCpG} shows the CpG sites selected at least 16 times in the top 20\% of the highly ranked CpG sites. The CpG sites FGF2\_P229\_F, IL1RN\_E42\_F, RARA\_P1076\_R, TFF1\_P180\_R, TGFB3\_E58\_R, WNT2\_P217\_F were consistently selected in the top 20\% of highly ranked CpG sites across all resampled datasets. \cref{fig:consistentMethAnovaSig} shows the relative methylation levels of the CpG sites that were consistently selected or were significantly different between those who survived and those who died. The mean methylation levels for the CpG sites IL1RN\_E42\_F and TGFB3\_E58\_R were statistically different between those who died and those who survived breast cancer (p-value $< 0.05$ from the Anova test). In particular, the mean relative methylation levels for IL1RN\_E42\_F and TGFB3\_E58\_R  were lower in those who died from breast cancer compared to those who did not. The interleukin 1 receptor antagonist (IL1RN) gene is a protein-coding gene that encodes the interleukin-1 receptor antagonist protein, a member of the interleukin 1 cytokine family.  
IL1RN is an anti-inflammatory molecule that modulates the biological activity of the pro-inflammatory cytokine, interleukin-1 \citep{shiiba2015interleukin}. IL1RN has been implicated in several cancers.

\begin{table}
% \spacingset{1}
\centering
	\begin{scriptsize}	
				\caption{Frequency of Genes selected at least 16 times in the top 20\% across 20 resampled datasets \label{tab: breastcancerVarSelCpG}}
    \resizebox{\textwidth}{!}{
				\begin{tabular}{llll}
			\hline
			\hline
			CpG Site& Corresponding Gene   & Gene Name & Frequency	\\
			\hline
			\hline
FGF2\_P229\_F&	FGF2&	fibroblast growth factor 2	&20\\
IL1RN\_E42\_F	&IL1RN	&interleukin 1 receptor antagonist&	20\\
RARA\_P1076\_R&	RARA&	retinoic acid receptor alpha&	20\\
TFF1\_P180\_R&	TFF1&	trefoil factor 1)&	20\\
TGFB3\_E58\_R	&TGFB3	&transforming growth factor beta 3	&20\\
WNT2\_P217\_F&	WNT2&	Wnt family member 2	&20\\
ADAMTS12\_E52\_R&	ADAMTS12&	ADAM metallopeptidase with thrombospondin type 1 motif 12&	19\\
RASSF1\_P244\_F&	RASSF1	&Ras association domain family member 1	&18\\
FABP3\_E113\_F&	FABP3	&fatty acid binding protein 3&	16\\
IGFBP7\_P297\_F&	IGFBP7	&insulin like growth factor binding protein 7&	16\\
IL1RN\_P93\_R	&IL1RN	&interleukin 1 receptor antagonist	&16\\
RASSF1\_E116\_F&	RASSF1&	Ras association domain family member 1&	16\\
SLC22A3\_E122\_R&	SLC22A3&	solute carrier family 22 member 3&	16\\
TPEF\_seq\_44\_S88\_R&	TPEF&	Transmembrane Protein With EGF Like & 16\\
~& ~&And Two Follistatin Like Domains 2&	~\\
			
\hline
			\hline
		\end{tabular}}
	\end{scriptsize}	
\end{table}	

%The methylation levels of 9 of the  CpG sites appearing at least 16 times were statistically different (p-value $< 0.05$ from Wilcoxon test) between survival status (dead or alive) or oestrogen receptor (ER) status (positive or negative) [Figures XXX and XXX]. The mean methylation levels 

% Figure \ref{fig:consistentMethAnovaSig} shows the CpG sites that were consistently selected in the top 20\% of highly-ranked CpG sites across all datasets.  The mean methylation levels for the CpG sites IL1RN\_E42\_F and TGFB3\_E58\_R were statistically different between those who died and those who survived from breast cancer (p-value $< 0.05$ from Anova test). 
\noindent{\textbf{\textit{Gene Ontology and Pathway Enrichment Analyses}}}:
We use an online enrichment tool, ToppGene Suite \citep{chen2009toppgene}, to explore the biological relationships of these ``stable" genes and CpG sites. We took these genes from the gene expression data and genes corresponding to the CpG sites as input for ToppFun in ToppGene Suite. Some of the biological processes enriched with gene ontology (GO) included vasculature development, tissue development, angiogenesis, and tube morphogenesis (see \cref{tab: GoBiologicalProcess,tab: GoBiologicalProcessCpGSites}). Some of the biological processes significantly enriched in our list of methylation include tissue morphogenesis, epithelial tube morphogenesis, and tube development. All these biological processes are essential in cell development, and aberrations or disruptions in these processes could result in cancer. \cref{tab: PathwaysCpGSites,tab: PathwaysGenes} show the top 10 pathways that are enriched in our list of methylated genes and genes, respectively. Some of these pathways included cancer and pathways related to extracellular matrix orgnaization (ECM). ECM is a complex collection of proteins and plays a key role in cell survival, cell proliferation, and differentiation \citep{naba2016extracellular}. ECM is involved in tumor progression, dissemination, and response to therapy \citep{naba2016extracellular,henke2020extracellular}.

\begin{table}
%\spacingset{1}
\centering
	\begin{scriptsize}	
				\caption{ Top 10 Gene Ontology (GO) Biological Processes enriched with ToppFun in ToppGene Suite\label{tab: GoBiologicalProcess}}
    \resizebox{\textwidth}{!}{
				\begin{tabular}{llll}
			\hline
			\hline
GO ID & GO Biological & Bonferroni & Genes \\
&Process& P-value & \\
\hline
GO:0001944&	vasculature development	& 2.71E-10&			DAB2,EPHB3,SLIT2,BIRC5,TCF4,THBS2,SPARC,MMP2\\
&&&TNFRSF1B,THY1,DCN,IGFBP7,PDGFRB,LOX,HLA-F,COL1A2\\							
	GO:0001568&	blood vessel development&	 2.10E-09&		DAB2,EPHB3,SLIT2,BIRC5,TCF4,THBS2,SPARC,MMP2\\
	&&&THY1,DCN,IGFBP7,PDGFRB,LOX,HLA-F,COL1A2	\\						
GO:0048514&	blood vessel morphogenesis&	 	9.21E-09&			DAB2,EPHB3,SLIT2,BIRC5,TCF4,THBS2,SPARC,MMP2\\
&&&THY1,DCN,IGFBP7,PDGFRB,LOX,HLA-F	\\						
	GO:0030198&	extracellular matrix organization&	 1.40E-08&		COL6A1,MFAP4,SPARC,MMP2,TNFRSF1B\\
	&&&DCN,TIMP2,LOX,VCAN,BGN,COL1A2\\							
	GO:0043062&	extracellular structure organization&	 	1.43E-08&		COL6A1,MFAP4,SPARC,MMP2,TNFRSF1B\\
	&&&DCN,TIMP2,LOX,VCAN,BGN,COL1A2			\\				
	GO:0045229&	external encapsulating structure organization&	 	1.53E-08&	COL6A1,MFAP4,SPARC,MMP2,TNFRSF1B\\
	&&&DCN,TIMP2,LOX,VCAN,BGN,COL1A2	\\						
GO:0072359&	circulatory system development&	 	1.83E-08&	DAB2,EPHB3,SLIT2,BIRC5,TCF4,THBS2,SPARC,MMP2,TNFRSF1B\\
&&&THY1,DCN,IGFBP7,PDGFRB,LOX,VCAN,HLA-F,COL1A2\\							
	GO:0001525&	angiogenesis&	 2.46E-08&		DAB2,EPHB3,SLIT2,BIRC5,TCF4,THBS2,SPARC\\
	&&&MMP2,THY1,DCN,IGFBP7,PDGFRB,HLA-F\\							
	GO:0035295&	tube development&	 1.39E-07&		DAB2,EPHB3,SLIT2,SPDEF,BIRC5,TCF4,THBS2,SPARC,MMP2\\
	&&&THY1,DCN,IGFBP7,PDGFRB,LOX,VCAN,HLA-F\\							
	GO:0035239&	tube morphogenesis&	 1.02E-06&	DAB2,EPHB3,SLIT2,BIRC5,TCF4,THBS2,SPARC,MMP2\\
	&&&THY1,DCN,IGFBP7,PDGFRB,LOX,HLA-F\\					
\hline
			\hline
		\end{tabular}}
	\end{scriptsize}	
\end{table}

\begin{table}
%\spacingset{1}
\centering
	\begin{scriptsize}	
				\caption{ Genes corresponding to CpG sites. Top 10 Gene Ontology (GO) Biological Processes  enriched with ToppFun in ToppGene Suite
    \label{tab: GoBiologicalProcessCpGSites}}
    \resizebox{\textwidth}{!}{
				\begin{tabular}{llll}
			\hline
			\hline
GO ID & GO Biological & Bonferroni & Genes \\
&Process& P-value & \\
\hline
GO:0048729&	tissue morphogenesis&	0.00003361&	ADAMTS12,IGFBP7,TGFB3,IL1RN,FGF2,WNT2,RARA,FABP3	\\
GO:0060562&	epithelial tube morphogenesis&	0.0002242&	ADAMTS12,IGFBP7,FGF2,WNT2,RARA,FABP3\\	
GO:0010092&	specification of animal organ identity&	0.003616&	FGF2,WNT2,RARA	\\
GO:0060591&	chondroblast differentiation&	0.004815&	FGF2,RARA	\\
GO:0008285&	negative regulation of cell population proliferation&	0.004981&	IGFBP7,TGFB3,FGF2,TFF1,RARA,FABP3\\	
GO:0002009&	morphogenesis of an epithelium&	0.005639&	ADAMTS12,IGFBP7,FGF2,WNT2,RARA,FABP3\\	
GO:0035295&	tube development&	0.01252&	ADAMTS12,IGFBP7,TGFB3,FGF2,WNT2,RARA,FABP3	\\
GO:0048598&	embryonic morphogenesis&	0.0138&	TGFB3,IL1RN,FGF2,WNT2,RARA,FABP3	\\
GO:0061035&	regulation of cartilage development&	0.01498&	ADAMTS12,TGFB3,RARA	\\
GO:1905330&	regulation of morphogenesis of an epithelium&	0.01603&	ADAMTS12,FGF2,WNT2	\\
\hline
			\hline
		\end{tabular}}
	\end{scriptsize}	
\end{table}

\begin{table}
\caption{Genes corresponding to CpG sites. Top 10 Pathways  enriched with ToppFun in ToppGene Suite.}
\label{tab: PathwaysCpGSites}
\begin{center}
\resizebox{\textwidth}{!}{
\begin{tabular}{lllll}
% \Hline
\hline
\hline
ID & Pathway & Source & Bonferroni & Genes \\
&~& ~&P-value & \\
\hline
M12868&	Pathways in cancer&	MSigDB C2 BIOCARTA &	0.001107&	TGFB3,FGF2,WNT2,RASSF1,RARA		\\
M39427&	Pluripotent stem cell differentiation pathway&	MSigDB C2 BIOCARTA&	0.002385&	TGFB3,FGF2,WNT2	\\
%~&	 pathway&	~&	~&	~	\\	
83105&	Pathways in cancer&	BioSystems: KEGG &	0.002876&	TGFB3,FGF2,WNT2,RASSF1,RARA		\\
M5889&	Ensemble of genes encoding extracellular  & 	MSigDB C2 BIOCARTA&	0.02151&	ADAMTS12,IGFBP7,TGFB3	\\	
~&	 matrix and extracellular & 	~& ~&	IL1RN,FGF2,WNT2	\\	
~&	matrix-associated proteins& 	~& ~&	~	\\	

M5883&	Genes encoding secreted soluble factors&	MSigDB C2 BIOCARTA &	0.04072&	TGFB3,IL1RN,FGF2,WNT2		\\
M5885&	Ensemble of genes encoding ECM-associated  &	MSigDB C2 BIOCARTA &	0.06319	&ADAMTS12,TGFB3,IL1RN,		\\
~&proteins including ECM-affilaited proteins, &	~&	~	&FGF2,WNT2	\\
~&ECM regulators&	~&	~	&~	\\
~&and secreted factors&	~&	~	&~	\\
138010&	Glypican 1 network&	BioSystems: Pathway Interaction &	0.06838&	TGFB3,FGF2	\\	
~&	~&	Database&~&	~	\\	
M33&	Glypican 1 network&	MSigDB C2 BIOCARTA&	0.07382&	TGFB3,FGF2		\\
749777&	Hippo signaling pathway&	BioSystems: KEGG&	0.07853	&TGFB3,WNT2,RASSF1	\\	
M12095&	Signal transduction through IL1R&	MSigDB C2 BIOCARTA&	0.09762&	TGFB3,IL1RN		\\
\hline
\hline
\end{tabular}}
\end{center}
\end{table}

%% Try end

% \begin{sidewaystable}
% \caption{Genes selected. Top 10 Pathways  enriched with ToppFun in ToppGene Suite.}
% \label{tab: PathwaysGenes}
% \begin{center}
% \resizebox{\textwidth}{!}{
% \begin{tabular}{lllll}
% \Hline
% ID & Pathway & Source & Bonferroni & Genes \\
% &~& ~&P-value & \\
% \hline
% 1270244	&Extracellular matrix organization&	BioSystems: REACTOME&	5.117E-08&	COL6A1,MFAP4,SPARC,MMP2\\
% ~&~&~&	~&	DCN,TIMP2,LOX,VCAN,BGN,COL1A2\\
% M5889&	Ensemble of genes encoding extracellular matrix &	MSigDB C2 BIOCARTA&	0.000000478&	SLIT2,COL6A1,MFAP4,THBS2\\
% ~&	and extracellular matrix-associated proteins &~&	~&IL16,SPARC,MMP2,DCN,IGFBP7\\
% ~&	~ &~&	~&ITIMP2,LOX,VCAN,BGN,COL1A2\\
% 1269016&	Defective CHSY1 causes TPBS&	BioSystems: REACTOME&	0.0001055&	DCN,VCAN,BGN\\
% 1269017	&Defective CHST3 causes SEDCJD&	BioSystems: REACTOME&	0.0001055&	DCN,VCAN,BGN\\
% 1269018	&Defective CHST14 causes EDS, &	BioSystems: REACTOME&	0.0001055&	DCN,VCAN,BGN\\
% ~&musculocontractural type&	~&	~&	~\\
% 1269986	&Dermatan sulfate biosynthesis&	BioSystems: REACTOME&	0.0004947&	DCN,VCAN,BGN\\
% 1269987	&CS/DS degradation&	BioSystems: REACTOME&	0.001087&	DCN,VCAN,BGN\\
% 1270256	&ECM proteoglycans&	BioSystems: REACTOME&	0.001966&	SPARC,DCN,VCAN,BGN\\
% 1309217	&Defective B3GALT6 causes EDSP2 and SEMDJL1&	BioSystems: REACTOME&	0.002875&	DCN,VCAN,BGN\\
% \hline
% \end{tabular}
% \end{center}
% \end{sidewaystable}

%% Han Lu try horizontal table 0818
\begin{table}
\caption{Genes selected. Top 10 Pathways  enriched with ToppFun in ToppGene Suite.}
\label{tab: PathwaysGenes}
\begin{center}
\resizebox{\textwidth}{!}{
\begin{tabular}{lllll}
% \Hline
\hline
\hline
ID & Pathway & Source & Bonferroni & Genes \\
&~& ~&P-value & \\
\hline
1270244	&Extracellular matrix organization&	BioSystems: REACTOME&	5.117E-08&	COL6A1,MFAP4,SPARC,MMP2\\
~&~&~&	~&	DCN,TIMP2,LOX,VCAN,BGN,COL1A2\\
M5889&	Ensemble of genes encoding extracellular matrix &	MSigDB C2 BIOCARTA&	0.000000478&	SLIT2,COL6A1,MFAP4,THBS2\\
~&	and extracellular matrix-associated proteins &~&	~&IL16,SPARC,MMP2,DCN,IGFBP7\\
~&	~ &~&	~&ITIMP2,LOX,VCAN,BGN,COL1A2\\
1269016&	Defective CHSY1 causes TPBS&	BioSystems: REACTOME&	0.0001055&	DCN,VCAN,BGN\\
1269017	&Defective CHST3 causes SEDCJD&	BioSystems: REACTOME&	0.0001055&	DCN,VCAN,BGN\\
1269018	&Defective CHST14 causes EDS, &	BioSystems: REACTOME&	0.0001055&	DCN,VCAN,BGN\\
~&musculocontractural type&	~&	~&	~\\
1269986	&Dermatan sulfate biosynthesis&	BioSystems: REACTOME&	0.0004947&	DCN,VCAN,BGN\\
1269987	&CS/DS degradation&	BioSystems: REACTOME&	0.001087&	DCN,VCAN,BGN\\
1270256	&ECM proteoglycans&	BioSystems: REACTOME&	0.001966&	SPARC,DCN,VCAN,BGN\\
1309217	&Defective B3GALT6 causes EDSP2 and SEMDJL1&	BioSystems: REACTOME&	0.002875&	DCN,VCAN,BGN\\
\hline
\hline
\end{tabular}}
\end{center}
\end{table}

\subsubsection{Goal 2: Model nonlinear relationships between methylation and gene expression data and derive molecular clusters}
%\noindent\normal{\textbf{\textit{Goal 2: Molecular Clustering}}}\\
We demonstrate the use of the estimated shared low-dimensional representation and the reconstructed methylation and gene expression data in molecular clustering. For this purpose, we applied the proposed method (without Laplacian) to all data to identify the top $20\%$ genes and CpG sites that could be used to nonlinearly approximate the original views. Then we obtained the shared low-dimensional representation ($\widetilde{\bZ}'$), and the reconstructed views ($R_1(\bZ')$, and $R_2(\bZ')$) based only on the top 20\% genes and CpG sites. We perform K-means clustering on $\widetilde{\bZ}'$,  $R_1(\bZ)$ and $R_2(\bZ)$. We set the number of clusters to 4, which is within the number of clusters investigated in the original article \citep{holm2010molecular}. We compared the number of clusters detected with several variables related to breast cancer, including estrogen receptor (ER) status, progesterone receptor (PgR) status, survival time, and survival status for ten years. We obtained Kaplan-Meier (KM) curves to compare the survival curves for the identified clusters. We also fitted a Cox regression model to compare the estimated hazard ratios for 10-year survival. Finally, we performed an enrichment analysis of the top 20\% genes and CpG sites. 

\begin{figure}
    \vspace{-1em}
    \centering
    \includegraphics[width=0.8\linewidth]{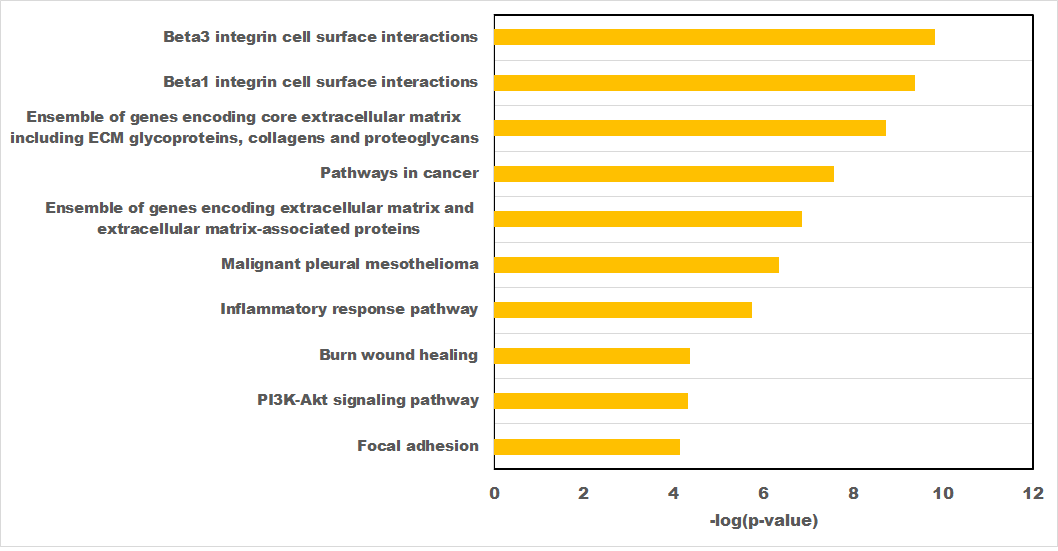} \includegraphics[width=0.8\linewidth]{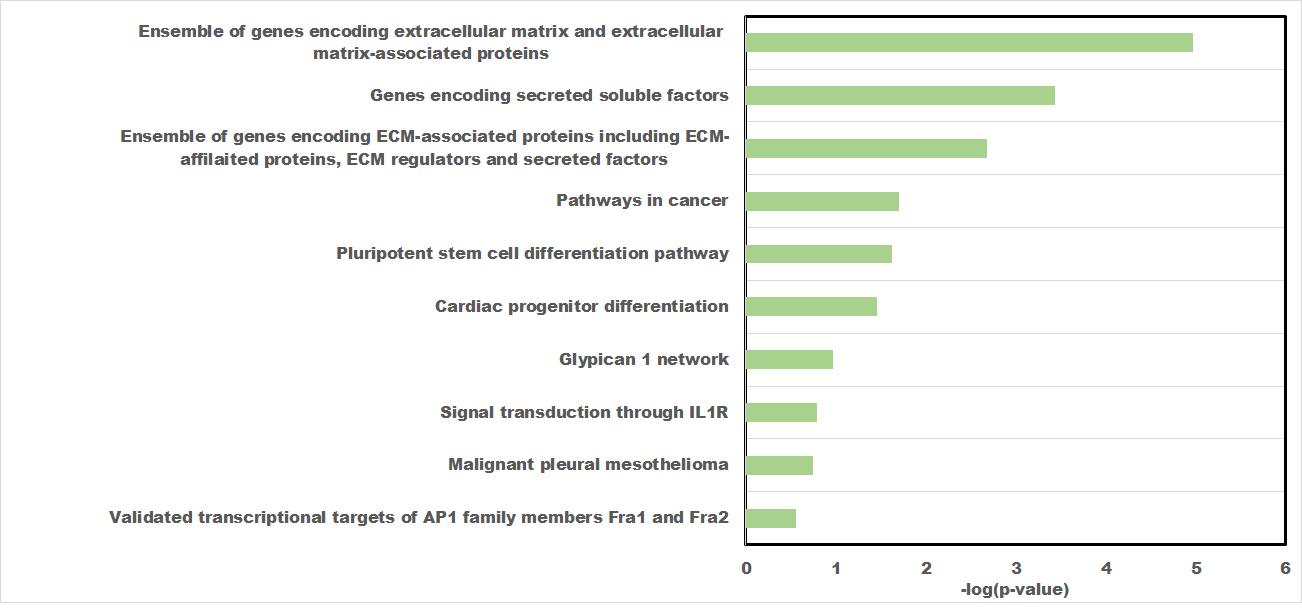} \\
    \caption{Top 10 significant pathways using highly-ranked genes (Top Panel) and genes corresponding to highly-ranked CpG sites (Bottom Panel)}
    \label{fig:clusterPathways}
\end{figure}

\begin{figure}
    \centering
    \includegraphics[width=0.8\linewidth]{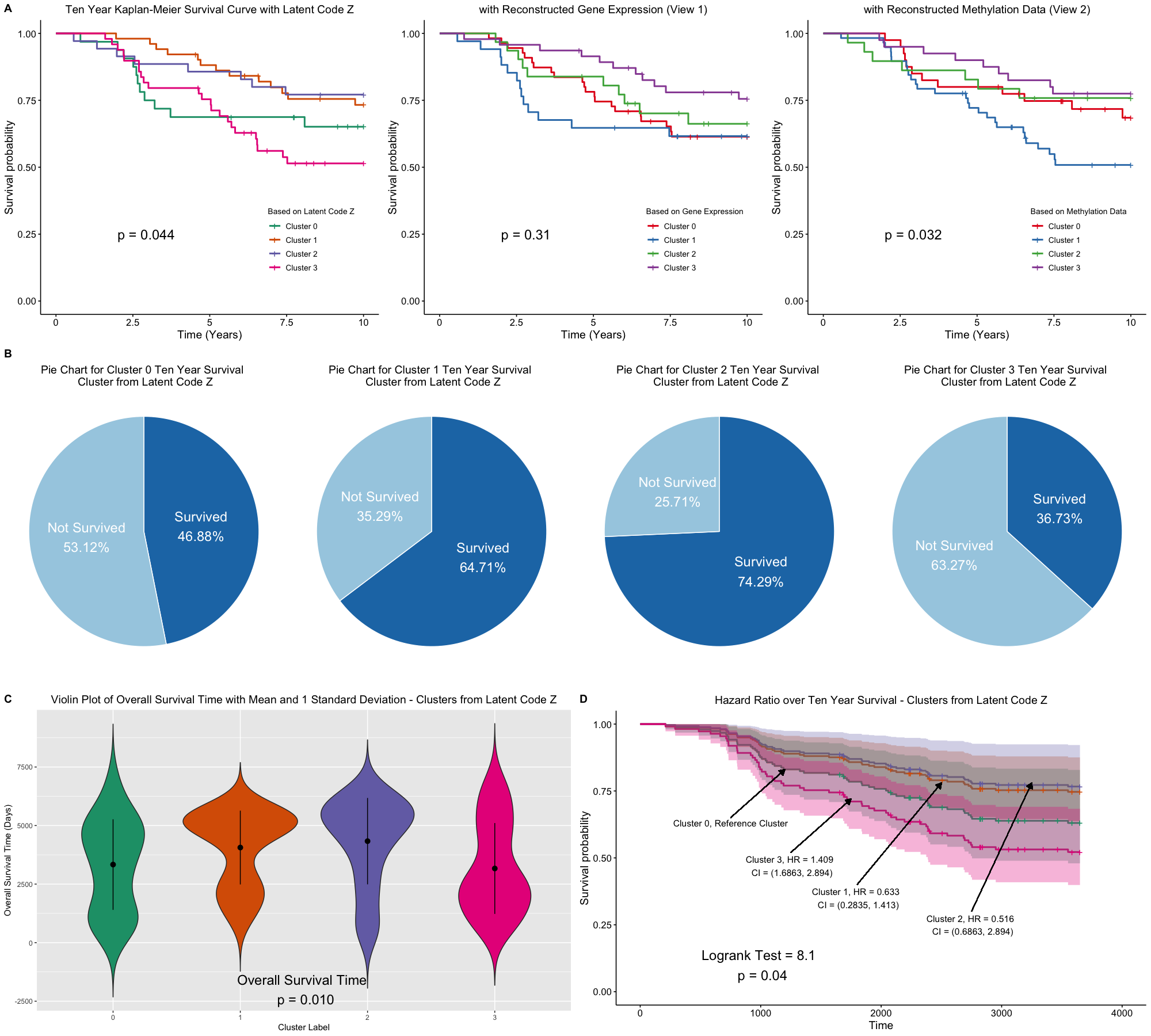}
    \caption{Top 20\% genes and CpG sites that approximate the original data are used to obtain shared low-dimensional representations, and reconstructed gene expression and methylation data. Figure A: Kaplan-Meier plots comparing survival curves for clusters obtained from the shared low-dimensional representations, and the reconstructed data. Survival curves for the  clusters based on the joint and low-dimensional representations and reconstructed methylation data are significantly different. Figures B-D: Clusters are derived from the shared low-dimensional representation. Figure B: Comparison of 10-year survival rates across clusters.  Chi-square test of independence shows that the clusters detected are significantly associated with 10-year survival event (p-value$=0.011$). Figure C: Violin plot of overall survival time by clusters. The average survival times are significantly different across clusters. D: Comparison of hazard ratios and survival curves across clusters. }
    \label{fig:survival}
\end{figure}

\cref{fig:survival} (A) shows the KM curves for the clusters detected using the low-dimensional shared representation (first panel) and the reconstructed gene expression (middle panel) and methylation data (right panel). From the KM plots, the 10-year survival curves for the clusters detected using the shared low-dimensional representation or the reconstructed methylation data are significantly different (p-value$=0.041$ and $0.032$, respectively, based on a log-rank test to compare survival curves). 
%Meanwhile, the survival curves for clusters detected from the reconstructed views are not any different. 
As reported in \cref{tab: patients}, the clusters (from shared low-dimensional representations) are significantly associated with ER, PgR, overall survival time, and 10-year survival event. Individuals in Cluster 3 seemed to have worse survival outcomes compared to individuals in Cluster 0. In particular, the proportion of individuals in Cluster 3 with ER/PgR negative tumors was higher, the 10-year survival rate was lower (only 40\% of participants from Cluster 0 survived while 69\% of participants from Cluster 1 survived, \cref{fig:survival} (B) ), and the average survival time was shorter compared to those in Cluster 1 (\cref{fig:survival} (C)). 
Furthermore, the estimated unadjusted risk ratio for 10-year survival for those in Cluster 3 compared to those in Cluster 0 was 1.409 (\cref{fig:survival} (D) 95\%CI: $1.686 - 2.894$, p-value$=0.04$), suggesting that being in Cluster 3 reduces your survival rate by a factor of 1.41 at each point during 10-year follow-up compared to Cluster 0. This effect persisted even after adjusting for age or age and ER status. Significantly enriched pathways, as shown in \cref{fig:clusterPathways}, from our gene list and CpG sites (genes corresponding to the top $20\%$ CpG sites) include ECM, inflammatory response pathway, and pathways in cancer. 

\begin{table}
% \centering

\caption{Characteristics of the patients. Continuous variables are tested based on regular ANOVA with equal variance assumption, and categorical variables are tested based on the Chi-square test.}
\label{tab: patients}
\begin{center}
\begin{scriptsize}
\resizebox{\textwidth}{!}{
\begin{tabular}{llllll}
\hline
\hline
  & Cluster 0 & Cluster 1 & Cluster 2 & Cluster 3 & p test\\
\hline
n & 32 & 51 & 35 & 50 &  \\
\hline
ER = er\_pos (\%) & 7 (22.6) & 37 (75.5) & 18 (51.4) & 43 (87.8) & $<$0.001 \\
\hline
PgR = pgr\_pos (\%) & 7 (22.6) & 38 (77.6) & 16 (45.7) & 39 (79.6) & $<$0.001 \\
\hline
Overall Survival & & & & & \\
~~~~Time (yr) (mean (SD)) & 9.14 (5.30) & 11.13 (4.32) & 11.87 (5.05) & 8.68 (5.31) & 0.010 \\
~~~~Event = 1 (\%) & 11 (34.4) & 18 (35.3) & 10 (28.6) & 26 (52.0) & 0.125 \\
\hline
Ten Year Survival & & & & & \\
~~~~Survived = 1 (\%) & 17 (53.1) & 18 (35.3) & 9 (25.7) & 31 (63.3) & 0.002 \\
\hline
HuSubtype (\%) &  &  &  &  & $<$0.001 \\
~~~~Basal & 21 (65.6) & 6 (11.8) & 12 (34.3) & 0 (0.0) &  \\
~~~~Her2 & 1 (3.1) & 5 (9.8) & 6 (17.1) & 2 (4.0) &  \\
~~~~LumA & 4 (12.5) & 15 (29.4) & 6 (17.1) & 19 (38.0) &  \\
~~~~LumB & 4 (12.5) & 8 (15.7) & 6 (17.1) & 14 (28.0) &  \\
~~~~non-classified & 0 (0.0) & 12 (23.5) & 4 (11.4) & 6 (12.0)  & \\
~~~~Normal & 2 (6.2) & 5 (9.8) & 1 (2.9) & 9 (18.0) &  \\
\hline
ageYear (mean (SD)) & 48.84 (9.64) & 52.16 (11.75) & 47.74 (10.46) & 49.80 (12.14) & 0.308 \\
\hline
\hline
\end{tabular}}
\end{scriptsize}
\end{center}
\end{table}

% \begin{figure}[H]
%     \centering
%     %\includegraphics[width=0.8\linewidth]{Plots/arranged_ggsave.png}
%     \includegraphics[width=0.98\linewidth]{Plots/PathwaysClusteringGenesTop20Percent2.png} \includegraphics[width=0.98\linewidth]{Plots/PathwaysClusteringMethTop20Percent.png} \\
%     \caption{Top 10 significant pathways using highly-ranked genes (Top Panel) and genes corresponding to highly-ranked CpG sites (Bottom Panel)}
%     \label{fig:clusterPathways}
% \end{figure}

% % ======== New Table for Real Data Summary 20231220 ========

% \begin{table}[]
%     \centering
%     \caption{Summary of real data, the datasets used and the tasks for each analysis.}
%     \resizebox{\textwidth}{!}{
%     \begin{tabular}{lllll}
%     \hline
%     \hline
%         Dataset & Categories & Number of features  & Sample Size & Task \\
%                 &            & in each view        &  & \\
%         \hline
%         Holm Breast Cancer & Died: 65 & View 1, gene expression, 469 & Training $n=112$ & Classification and  \\
%         Study              & Survived: 103 & View 2, methylation, 334     & Testing $n=56$  & Clustering\\
%         \hline
%         Shear Transformed & Hand-written digits 0 to 9 & View 1, digits, 784 & Training $n=60000$ & Classification and  \\
%         MNIST Dataset & count ranging from 5400 to 6800 & View 2, digits, 784 & Testing $n=10000$ & Reconstruction\\
%         \hline
%         LGG Dataset & Grade 2: 246 & View 1, methylation, 9691 & Training $n=410$ & Classification \\
%                     & Grade 3: 264 & View 2, miRNA, 235        & Testing $n=100$  & \\
%                     &              & View 3, mRNAseq, 7603 & & \\
%     \hline
%     \hline
%     \end{tabular}
%     }
%     \label{tab:my_label}
% \end{table}

\subsection{{Evaluation of Brain Lower Grade Glioma Data}}
\label{s: real_lgg}
%Description of LGG.
We applied our method to data pertaining to brain lower grade glioma (LGG) to identify molecules that discriminate between levels of LGG grade (grade $2$ vs $3$ gliomas). We obtained data from the Board GDAC Firehose of the Cancer Genome Atlas Program (TCGA)\footnote{\url{https://gdac.broadinstitute.org}}. We used three types of omics data: methylation, miRNA, and mRNAseq, following the analysis in \cite{wang_mogonet_2021}. Only patients with all available omics and classifications of grade were included in our analyzes, giving a total sample size of $510$, with $246$ patients classified as grade $2$ and $264$ patients as grade $3$.
%Goal of analysis.
We used the LGG dataset to demonstrate that the proposed method can be used to associate three views, select important biomarkers, and make accurate predictions of the patient grade category. 

%Data preprocessing and data cleaning.
Data cleaning and data preprocessing were carried out on each view of data to remove features with low potential for discrimination. For all views, we first removed features with missing measures. Due to the limited number of features left in the miRNA view after removing missing values, future preprocessing was conducted only on the DNA methylation view and the mRNAseq view. Unsupervised preprocessing was applied to remove features whose variance was less than $0.001$ for DNA methylation measures and $0.1$ for mRNAseq measures, following the thresholds used in \cite{wang_mogonet_2021}. The data were then divided into training sets ($n=410$) and testing ($n=100$) sets and supervised preprocessing was conducted on the training set. Logistic regression was fitted for each feature in the DNA methylation view and the mRNAseq view. The p-values were adjusted by the Benjamini-Hochberg procedure, and the features with adjusted p-values $<0.05$ were kept in the dataset. After data cleaning and preprocessing, the number of features for DNA methylation, miRNA, and mRNAseq was $9691$, $235$, and $7603$ respectively.

%Method compared with and citation. 
We applied the proposed approach to the training dataset, where we selected the important features from each type of omics data. Subsequently, we used these selected features to make predictions for the patient's grade category in the testing dataset, as shown in \cref{tab: lgg}. We used cross-validation to tune hyper-parameters based on the training set. Our proposed method was compared with Deep Generalized Canonical Correlation Analysis (Deep GCCA) \cite{benton_deep_2017} with PyTorch implementation \footnote{\url{https://github.com/arminarj/DeepGCCA-pytorch}}. We add the teacher-student network (TS) \cite{TS:2019} for feature selection, and implement SVM for classification; Deep IDA \cite{wang2021deep}; Features selected from Deep IDA with SVM for classification; SIDA \cite{SIDA:2019}, and SVM and Random Forest on stacked data. The classification performance is presented in \cref{tab: lgg}.

%Comparison Result
\begin{table}
\caption{LGG dataset: SVM and random forest are based on stacked views. Deep IDA + SVM means selecting features from Deep IDA and training an SVM classifier on these features. iDeepViewLearn with selected top $50$ features obtains a classification error based on a shared low-dimensional representation trained on data with the selected top $50$ features. Similar for iDeepViewLearn with selected top $100$ features.}
\label{tab: lgg}
\begin{center}
\begin{tabular}{lr}
% \Hline
Method&AverageError (\%)  	\\
			\hline
			\hline
    			SVM on stacked data & 30.00  \\
                Random Forest on stacked data & 26.00  \\
    			\textbf{iDeepViewLearn} with selected top 50 features & 28.00\\
    			\textbf{iDeepViewLearn} with selected top 100 features & 26.00\\
                % \textbf{iDeepViewLearn} with selected top 200 features & 26.00\\
                SIDA & 29.00 \\
                Deep GCCA + SVM & 29.00 \\
                Deep IDA & 28.00\\
                Deep IDA + SVM & 26.00 \\
    			
    			%\textbf{Deep IDA + NCC + Bootstrap} & 76.45\\
%                \textbf{RKCCA} (Need to run this) & 42.41 (1.04)\\
\hline
\end{tabular}
\end{center}
\end{table}

\begin{figure}
    \centering
    \includegraphics[width=1\linewidth]{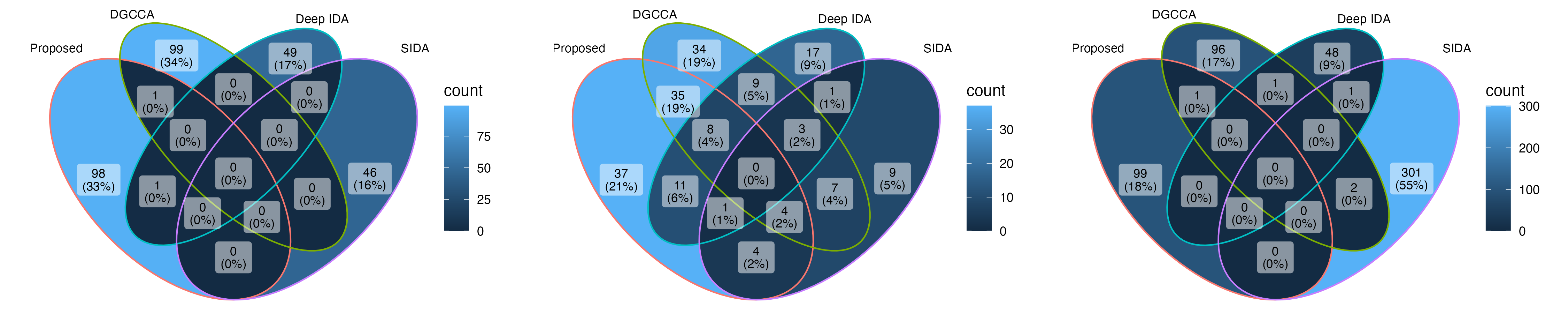}
    \caption{Venn diagrams of features selected by the proposed method, and the three comparison methods that conducted feature selection. The left, middle, and right panels correspond to the DNA methylation, miRNA, and mRNAseq view, respectively. The percentages represent the proportion of the total selected features from the four methods.}
    \label{fig:venn}
\end{figure}

%Feature Selection
In Figure\cref{fig:venn}, we show the overlaps of features selected by the methods. We used the top 100 features of each view selected by the proposed method. We compare the top 100 features selected by the TS network with Deep GCCA and the top 50 features selected by the TS network with Deep IDA. SIDA selected 46, 29, and 304 features for each omics, respectively. We presented the overlaps between the selected genes across the four methods matched from NCBI\footnote{\url{https://www.ncbi.nlm.nih.gov/}}. The overlaps between $2$ or more methods of DNA methylation were COL11A2 and FBLN2. The overlaps between $3$ or more methods for miRNA view were MIR379, MIR409, MIR29C, MIR129-1, MIR20B, MIR30E, MIR92A2, MIR222, MIR24-2, MIR767, MIR128-2, MIR105-2, and MIR17. The overlaps between 2 or more methods for mRNAseq view were NCAPH, LY86-AS1, HSFX2, and SLC25A41.

% % ==================== MNIST ====================
\subsection{{Evaluation of Shear Transformed MNIST Data}}
\label{s: real_mnist}

\begin{figure}
    \centering
    \includegraphics[width=0.98\linewidth]{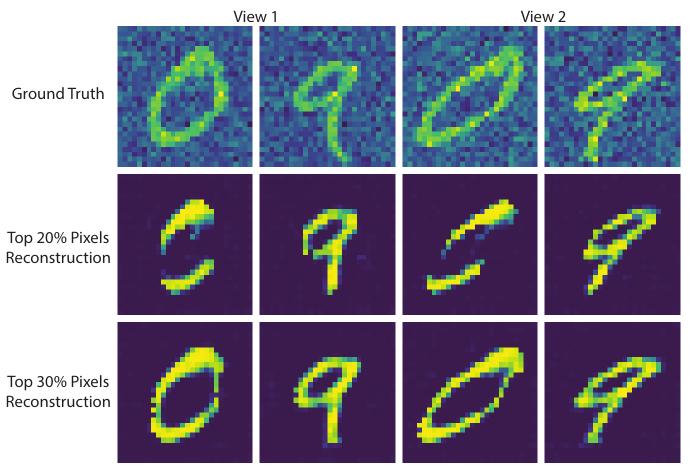}
    \caption{An example of shear transformed MNIST dataset. For the subject with label "0" and "9", view 1 observation is on the left and view 2 observation is on the right. Notably, we show the grayscale images with color only for better visualization.}
    \label{fig:mnist}
\end{figure}

We apply our method to the MNIST dataset \citep{lecun_gradient-based_1998}. The MNIST handwritten image dataset consists of 70,000 images of handwritten digits divided into training and testing sets of 60,000, and 10,000 images, respectively. 
%The MNIST database of handwritten digits has a training set of 60,000 examples, and a test set of 10,000 examples. 
% It is a subset of a larger set available from MNIST. 
The digits have been size-normalized and centered in a fixed-size image. Each image is $28 \times 28$ pixels and has an associated label that denotes which digit the image represents (0-9). We make good use of a shear mapping to generate a second view of these handwritten digits. A shear mapping is a linear map that displaces each point in a fixed direction by an amount proportional to its signed distance from the line that is parallel to that direction and goes through the origin. \cref{fig:mnist} shows two image plots of a digit for views 1 and 2.  

We used the MNIST dataset to demonstrate the ability of the proposed method to reconstruct handwritten images using a few pixels. In particular, neural networks $G_d$ consist of convolutional layers instead of fully connected layers, since they reconstruct images. We apply the proposed method to the training dataset, select $20\%$ and $30\%$ of the pixels based on our variable ranking criteria and reconstruct the images using only the selected pixels. We also learn a new model with these pixels, we use the learned model and the testing data to classify the test digits, and we obtain the test errors. \cref{fig:mnist} shows the reconstructed images based on the top $20\%$ and $30\%$ pixels. The digits are apparent even with only $30\%$ of the pixels. From \cref{tab: mnist}, the classification performance using the top $30\%$ of the pixels is comparable to Deep CCA and SVM, which use all pixels. Even when only $20\%$ of the pixels were selected and used to reconstruct the images, the classification performance of our method was competitive.

\begin{table}
\caption{MNIST dataset: SVM is based on stacked views. Deep CCA + SVM is a training SVM based on the last layer of Deep CCA. iDeepViewLearn with selected top $20\%$ pixels obtains a classification error based on a shared low-dimensional representation trained on data with the selected $20\%$ of the pixels. Similar for iDeepViewLearn with selected top $30\%$.}
\label{tab: mnist}
\begin{center}
\begin{tabular}{lr}
% \Hline
Method&AverageError (\%)  	\\
			\hline
			\hline
			    %All Random Forest Results - turns out similar to SVM
			    %Should not set max_depth
    			%Random Forest by itself & 32.14 (updated 0614) \\
    			%Deep CCA + Random Forest & 36.37 (0614 full batch) \\
    			%Deep CCA (shallow) + Random Forest & 24.66 (0621 batch size 1000) \\
    			%Deep CCA (shallow) + Random Forest & 28.91 (0621 batch size 5000) \\
    			%Deep CCA (shallow) + Random Forest & 32.12 (0621 batch size 10000) \\
    			%Deep CCA (shallow) + Random Forest & 33.48 (0614 full batch) \\
    			
    			% Han: the most recent error rate from my end seems to be 2.97 instead of 2.80...?
    			Deep CCA + SVM & 2.97  \\
    			%Deep CCA  + SVM & 2.80  \\
    			% Han: which is the same date on getting this 2.81. 
    			SVM on stacked data & 2.81  \\
    % 			Sparse CCA + Random Forest & 71.84 (0607 default 0612 still running) \\
    % 			\textbf{iDeepViewLearn} on selected top 10\% features & 8.81\\
    % 			\textbf{iDeepViewLearn} on selected top 20\% features & 5.01\\
    % 			\textbf{iDeepViewLearn} on selected top 30\% features & 3.51\\
    			\textbf{iDeepViewLearn} with selected top 20\% pixels & 3.91\\
    			\textbf{iDeepViewLearn} with selected top 30\% pixels & \textbf{2.56}\\
    			
    			%\textbf{Deep IDA + NCC + Bootstrap} & 76.45\\
%                \textbf{RKCCA} (Need to run this) & 42.41 (1.04)\\
\hline
\end{tabular}
\end{center}
\end{table}

% ==================== DISCUSSION ====================

\section{Discussion}
\label{s:Disc}

We have presented iDeepViewLearn, short for Interpretable Deep Learning Method for Multiview Learning, to learn nonlinear relationships in data from multiple sources. iDeepViewLearn combines the flexibility of deep learning with the statistical advantages of data- and knowledge-driven feature selection to yield interpretable results. In particular, iDeepViewLearn learns low-dimensional representations of the views that are common to all the views and assumes that each view can be approximated by a nonlinear function of the shared representations. Deep neural networks are used to model the nonlinear function and an optimization problem that minimizes the difference between the observed data and the nonlinearly transformed data are used to reconstruct the original data. A regularization penalty is imposed on the reconstructed data in the optimization problem, permitting us to reconstruct each view only with relevant variables. Beyond the data-driven approach for feature selection, we also consider a knowledge-based approach to identify relevant features. We use the normalized Laplacian of a graph to model bilateral relationships between variables in each view and to encourage the selection of connected variables.   

We have developed a user-friendly algorithm in Python 3, specifically PyTorch, and interfaced it with R to increase the reach of our method.  
%The algorithm for iDeepViewLearn, developed in Python 3 (specifically Pytorch), is user-friendly and will prove useful in many  data integration applications. 
Extensive simulations with varying data dimensions and complexity revealed that iDeepViewLearn outperforms several other linear and nonlinear methods for integrating data from multiple views, even in high-dimensional scenarios where the sample size is typically smaller than the number of variables.    

When iDeepViewLearn was applied to methylation and gene expression data related to breast cancer, we observed that iDeepViewLearn is capable of achieving meaningful biological insights. We identified several CpG sites and genes that better discriminated people who died from breast cancer and those who did not. The biological processes of the gene ontology enriched in the top-ranked genes and methylated CpG sites included processes essential to cell proliferation and death. The enriched pathways included cancer and others that have been implicated in tumor progression and response to therapy. Using the shared low-dimensional representations of gene expression and methylation data from our method, we detected four molecular clusters that differed in their 10-year survival rates. The enrichment analysis of highly ranked genes and genes corresponding to the CpG sites selected by our method showed a strong enrichment of pathways and biological processes, some related to breast cancer and others that could be further explored for their potential role in breast cancer. We also applied iDeepViewLearn to DNA methylation, miRNA, and mRNASeq data pertaining to Brain Lower Grade Glioma (LGG) and found our method to be competitive in discriminating between LGG categories, demonstrating the ability of our methods to be used for more than two views. We further applied iDeepViewLearn to handwritten image data and we were able to reconstruct the digits with about $30\%$ pixels while also achieving competitive classification accuracy. A limitation of our work is that the number (or proportion) of top-ranked features needs to be specified in advance. 

% The top disease and disorders significantly enriched in our list of genes, proteins, and metabolomics data included cancer, neurological disorders, infectious diseases, and metabolic diseases.  While some of these findings corroborate earlier results, the top-ranked molecules could be further investigated to delineate their impact on COVID-19 status and severity. 

%Our work has some limitations. First, the number (or proportion) of top-ranked features need to be specified in advance.  Second, XXX

\section{Conclusion}
\label{s:Conc}

In conclusion, we have developed deep learning methods to learn nonlinear relationships in multiview data that are able to identify features likely driving the overall association in the views. The simulations and real data applications are encouraging, even for scenarios with small to moderate sample sizes, thus we believe the methods will motivate other applications.

\backmatter

\section*{Declarations}

\subsection*{Ethics approval and consent to participate}
Not Applicable. This research uses publicly available data. 

\subsection*{Consent for publication}
Not applicable.

\subsection*{Availability of data and materials}
The data used were obtained from \cite{holm2010molecular}. 
We provide  Python codes and an R package, \textit{iDeepViewLearn}, to facilitate the use of our method. Its source codes, along with a README file, are available at: \url{https://github.com/lasandrall/iDeepViewLearn}. 

\subsection*{Competing interests}
The authors declare that they have no competing interests.

\subsection*{Funding}
The project described was supported by the Award Number 1R35GM142695 of the National Institute of General Medical Sciences of the National Institutes of Health. The content is solely the responsibility of the authors and does not represent the official views of the National Institutes of Health.. 

\subsection*{Authors' contributions}
SES and JS conceived of the idea and developed the methods. HW developed algorithms to implement the methods. HW and HL conducted simulations and real data analyses. SES and HW wrote a first draft of the paper. All authors edited and read the final manuscript.

% %%=============================================%%
% %% For submissions to Nature Portfolio Journals %%
% %% please use the heading ``Extended Data''.   %%
% %%=============================================%%

% %%=============================================================%%
% %% Sample for another appendix section			       %%
% %%=============================================================%%

% %% \section{Example of another appendix section}\label{secA2}%
% %% Appendices may be used for helpful, supporting or essential material that would otherwise 
% %% clutter, break up or be distracting to the text. Appendices can consist of sections, figures, 
% %% tables and equations etc.

% \end{appendices}

%%===========================================================================================%%
%% If you are submitting to one of the Nature Portfolio journals, using the eJP submission   %%
%% system, please include the references within the manuscript file itself. You may do this  %%
%% by copying the reference list from your .bbl file, paste it into the main manuscript .tex %%
%% file, and delete the associated \verb+\bibliography+ commands.                            %%
%%===========================================================================================%%

\bibliography{sn-bibliography}% common bib file
%% if required, the content of .bbl file can be included here once bbl is generated
%%\input sn-article.bbl

\end{document}

\renewcommand{\figurename}{Fig.}
\renewcommand{\thefigure}{S\arabic{figure}}

\renewcommand{\tablename}{Tab.}
\renewcommand{\thetable}{S\arabic{table}}
\title[iDeepViewLearn for Multiview Learning]{Supplementary Materials of Interpretable Deep Learning Methods for Multiview Learning}

%%=============================================================%%
%% Prefix	-> \pfx{Dr}
%% GivenName	-> \fnm{Joergen W.}
%% Particle	-> \spfx{van der} -> surname prefix
%% FamilyName	-> \sur{Ploeg}
%% Suffix	-> \sfx{IV}
%% NatureName	-> \tanm{Poet Laureate} -> Title after name
%% Degrees	-> \dgr{MSc, PhD}
%% \author*[1,2]{\pfx{Dr} \fnm{Joergen W.} \spfx{van der} \sur{Ploeg} \sfx{IV} \tanm{Poet Laureate} 
%%                 \dgr{MSc, PhD}}\email{iauthor@gmail.com}
%%=============================================================%%

\author[1]{\fnm{Hengkang} \sur{Wang}}\email{wang9881@umn.edu}

\author[2]{\fnm{Han} \sur{Lu}}\email{lu000054@umn.edu}

\author[1]{\fnm{Ju} \sur{Sun}}\email{jusun@umn.edu}

\author*[2]{\fnm{Sandra E} \sur{Safo}}\email{ssafo@umn.edu}

\affil[1]{\orgdiv{Department of Computer Science and Engineering}, \orgname{University of Minnesota}, \orgaddress{\city{Minneapolis}, \postcode{55455}, \country{USA}}}

\affil*[2]{\orgdiv{Division of Biostatistics}, \orgname{University of Minnesota}, \orgaddress{\city{Minneapolis}, \postcode{55455}, \country{USA}}}

\maketitle

\section{More on Methods}
\subsection{Optimization and Algorithm }
We solve the optimization problems iteratively using ADAM\citep{kingma_adam_2017} with $\beta_1 = 0.9$ and $\beta_2 = 0.999$. We initialize the weights and biases of the network by using the default settings in PyTorch and the shared low-dimensional representation matrix with numbers drawn from the standard distribution. Our algorithm is divided into three stages. The  first stage is the Feature Selection stage. In this stage, we solve the optimization problem (3) or (5) [main text] to obtain features that are highly-ranked. In particular, we feed forward the network with the weights and biases to obtain $G_d(\bZ)$. We compare $G_d(\bZ)$ with the training data and measure the error using the loss function. We perform a backwards pass and propagate the error to each individual node using backpropagation.  We compute gradients of the loss function with respect to the inputs of iDeepViewLearn (i.e. with respect to the weights and biases), for $\bZ$ fixed, and adjust weights using gradient descent. Then, we optimize the loss function with respect to $\bZ$, while keeping the weights and biases fixed. We repeat the process iteratively until the difference between the current and previous iteration losses is very small or a maximum number of iterations reached. Please refer to Algorithm 1 for more details.   

The second stage is the Reconstruction and Training stage using selected features. Here, we solve the optimization problem (6) [main text]. Our input data are the observed data with the selected features (i.e. top $r$ or $r\%$ features in each view), $\bX^{'(1)}\ldots \bX^{'(D)}$. We proceed similarly like the first stage, and we iteratively solve the optimization problem until convergence or some maximum number of iteration reached. At convergence, we obtain the reconstructed data $R_d(\bZ')$, and the learned shared low-dimensional representations, $\widetilde{\bZ'}$ based on only the top $r$ or $r\%$ variables in each view. Downstream analyses such classification, regression, or clustering could be carried out on this learned shared low-dimensional representations. For instance, for our simulations and real data analyses, we trained a support vector machine classifier using these shared low-dimensional representations. Additionally, for the real data analyses, we trained a K-means clustering algorithm on the shared low-dimensional representation to obtain clusters common to all the views. We note that downstream analyses such as clustering could be implemented on the reconstructed views for view-specific clusters, in addition to the joint clusters. Please refer to Algorithm 2 for more details.  

The third stage is the Prediction stage, if an outcome is available. Here, we solve the optimization problem (8) for the learned shared-low dimensional representation ($\widetilde{\bZ}'_{test}$) corresponding to the testing views ($\bX^{'(1)}_{test} \ldots \bX^{'(D)}_{test}$). We use the learned weights and biases from the second stage, and we implement a stochastic gradient descent algorithm. Since the network parameters (i.e. weights and biases) are fixed, we only update the loss  function with respect to $\bZ'_{test}$. Please refer to Algorithm 3 for more details. We use the automatic differentiation function ``autograd" in Pytorch to estimate the gradients of our loss function with respect to the network parameters or latent code. We use Exponential Linear Unit (ELU) \citep{clevert_fast_2016} as our nonlinear activation function. 

\subsection{Hyper-Parameter Selection}
There are several crucial hyper-parameters for our proposed model, including $\lambda^d$, $\lambda^{'d}$, number of latent components $K$, learning rates of neural networks for various views, and learning rate of the latent code $\bf Z$. We also add essential hyper-parameters of downstream models to our search space. For example, we also tune regularization parameter $C$ and kernel coefficient $\gamma$ when adopting Support Vector Machines (SVMs) with Radial Basis Function (RBF) for classification tasks. Particularly, $\lambda^d$ is set to be $0.1$ as default due to its insensitivity; the upper bound of $K$ is $r$ or the minimum of the numbers of features from all the views times $r\%$; learning rates for different neural networks are unified for simplicity. Since the hyper-parameter search space is not small, we obtain all combinations of our hyper-parameters, and we randomly select \citep{bergstra2012random} some combinations of hyper-parameters to search over. 
\begin{algorithm}
    \label{alg:fs}
    \caption{First stage: Feature Selection}
  \begin{algorithmic}[1]
  
    \INPUT Training data from different views $\bX^{(1)}, \bX^{(2)}, ..., \bX^{(D)}$. Each variable in a view is standardized to have mean zero and variance one, but no standardization is allowed when facing images because variables, i.e., pixels are not independent;  neural networks $G_1, G_2, ..., G_D$ for each view; Laplacian ($\mathcal{L}^{(d)}$) for network-based approach; 
    learning rate $l_1$, $l_2$, ..., $l_D$ for neural networks $G_1$, $G_2$, ..., $G_D$ respectively; latent code (or shared low-dimensional representation) $\bZ$ with $K$ components; learning rate $l_Z$ for the latent code $\bZ$; number of iterations $I$
    \OUTPUT Optimized weights and biases for neural networks $G_1, G_2, ..., G_D$; optimized latent code $\bZ$; indices $\bI_1$, $\bI_2$, ...,$\bI_D$ for important features of each view
    \STATE \textbf{Initialization} Initialize neural networks $G_1, G_2, ..., G_D$ by the default settings of $PyTorch$; Assign random numbers from the standard normal distribution for the latent code $\bZ$
    \WHILE{$i = 1, 2, ..., I$}
      \STATE Feed forward the network of latest weights and biases to obtain the reconstructions of each view (i.e. $G_d(\bZ)$)
      \STATE Apply equation (3) or (5) to calculate losses with $\bX^{(1)}$, \ldots,$\bX^{(D)}$
      \STATE Compute the gradient of weights and biases for each network and the latend code $\bZ$ by the $PyTorch$ $  Autograd$ function
      \STATE Update the weights and biases with specified learning rate $\alpha$
    \ENDWHILE
    \STATE
    \textbf{Feature Selection} Feed the learned latent code $\bZ$ into neural networks $G_1$, $G_2$, ..., $G_D$ to obtain reconstructed data $G_d(\bZ)$ for the observed data $\bX^{(d)}$. Calculate column-wise norm of $G_d(\bZ)$.  Choose the columns with large column norm as important features for that view. Save the indices of important features as $\bI_1$, $\bI_2$, ...,$\bI_D$. Denote the observed datasets with only the selected features as 
    $\bX^{\prime (1)}, \bX^{\prime (2)}, ..., \bX^{\prime (D)}$. 
  \end{algorithmic}
\end{algorithm}

% \begin{figure}[!htbp]
%     \centering
%     \includegraphics[width=1.0\linewidth]{Plots/fs.png}
%     \caption{Feature Selection. We train a deep learning model that takes all the views, estimates a shared low-dimensional representation $\bZ$ and obtains nonlinear reconstructions ($G_1(\bZ)$,\ldots,$G_D(\bZ)$) of the original views. We impose sparsity constraints on the reconstructions allowing us to identify a subset of variables for each view ($\bI_1$, \ldots,$\bI_D$) that  approximate the original data.}
%     \label{fig:fs}
% \end{figure}

\begin{algorithm}
\label{alg:rec}
    \caption{Second stage: Reconstruction and Training}
  \begin{algorithmic}[2]
    \INPUT Training data from different view $\bX^{(1)}, \bX^{(2)}, ..., \bX^{(D)}$, optional class labels or outcome $\mathbf{y}$;  neural networks $R_1, R_2, ..., R_D$ for each view; number of components $K$ for  latent code $\bZ^{\prime}$; learning rate $l_1^{\prime}$, $l_2^{\prime}$, ..., $l_D^{\prime}$ for neural networks $R_1$, $R_2$, ..., $R_D$ respectively; learning rate $l_Z^{\prime}$ for the latent code $\bZ^{\prime}$; number of iterations $I$; indices $\bI_1$, $\bI_2$, ...,$\bI_D$ for selected features for views $1,\ldots,D$, respectively.
    \OUTPUT Optimized weights and biases for neural networks $R_1, R_2, ..., R_D$; Optimized latent code $\bZ^{\prime}$ (denoted as $\widetilde{\bZ}^{\prime}$); a trained classifier $\textbf{C}$ or prediction model
    \STATE \textbf{Initialization} Initialize neural networks $R_1, R_2, ..., R_D$ by the default settings of $PyTorch$; Assign random numbers from the standard normal distribution for the latent code $\bZ^{\prime}$; Obtain $\bX^{\prime (1)}, \bX^{\prime (2)}, ..., \bX^{\prime (D)}$ by applying feature indices $\bI_1$, $\bI_2$, ...,$\bI_D$ to training data
    \WHILE{$i = 1, 2, ..., I$}
      \STATE Feed forward the network of latest weights and biases to obtain the reconstructions of each view (i.e. $R_d(\bZ^{\prime})$)
      \STATE Apply equation (7) to calculate losses with $\bX^{\prime (1)}, \bX^{\prime (2)}, ..., \bX^{\prime (D)}$
      \STATE Compute the gradient of weights and biases for each network and the latent code $Z$ by the $PyTorch$ $  Autograd$ function
      \STATE Update the weights and biases with specified learning rate $\alpha$
    \ENDWHILE
    \STATE
    \textbf{Training} Train a classifier $\textbf{C}$ by using the trained latent code $\widetilde{\bZ}^{\prime}$ and outcome $\mathbf{y}$
  \end{algorithmic}
\end{algorithm}

% \begin{figure}[!htbp]
%     \centering
%     \includegraphics[width=1.0\linewidth]{Plots/train.png}
%     \caption{Reconstruction and Downstream Analysis. We train a deep learning model to obtain a common low-dimensional representation $\bZ'$ that is based on the features selected in Algorithm 1, we obtain nonlinear approximations ($\bR_1(\bZ)$,\ldots,$\bR_D(\bZ))$, and we perform downstream analyses using estimated $\bZ'$.}
%     \label{fig:train}
% \end{figure}

\begin{algorithm}
\label{alg:test}
    \caption{Last stage: Testing}
  \begin{algorithmic}[3]
    \INPUT Testing data from different view $\bX^{(1)}_{test}, \bX^{(2)}_{test}, ..., \bX^{(D)}_{test}$;  learned neural networks $R_1, R_2, ..., R_D$  (i.e. weights and biases) for each view; number of components $K$ for  latent code $\bZ^{\prime}_{test}$; learning rate $l_{test}$ for the latent code $\bZ^{\prime}_{test}$; number of iterations $I$; indexes $\bI_1$, $\bI_2$, ...,$\bI_D$ for selected features for views $1,\ldots,D$, respectively; (optional) trained classifier ($\mathbf{C}$). 
    \OUTPUT Optimized latent code ${\bZ}^{\prime}_{test}$ of testing data (denoted as $\widetilde{\bZ}^{\prime}_{test})$; predicted outcome  $\widehat{\mathbf{y}}_{test}$
    \STATE \textbf{Initialization} Fix neural networks $R_1, R_2, ..., R_D$ trained in the second stage; Assign random numbers to latent code ${\bZ}_{test}$; Obtain ${\bX}_{test}^{\prime (1)}, {\bX}_{test}^{\prime (2)}, ..., {\bX}_{test}^{\prime (D)}$ by applying important features $\bI_1$, $\bI_2$, ...,$\bI_D$ to testing data
    \WHILE{$i = 1, 2, ..., I$}
      \STATE Feed forward the network by the latest ${\bZ}_{test}$ to obtain the reconstructions of each view (i.e $R_d(\bZ'_{test})$)
      \STATE Apply equation (8) to calculate losses with ${\bX}_{test}^{\prime (1)}, {\bX}_{test}^{\prime (2)}, ..., {\bX}_{test}^{\prime (D)}$
      \STATE Update ${\bZ'}_{test}$ with specified learning rate $l_{test}$
    \ENDWHILE
    \STATE
    \textbf{Testing} Put learned $\widetilde{\bZ'}_{test}$ into the trained classifier $\textbf{C}$ and obtain the predicted outcome $\widehat{\mathbf{y}}_{test}$
  \end{algorithmic}
\end{algorithm}
% \subsection{Unique Features of Proposed Method}
% \begin{table} [H]
% \label{tab:uniquefeatures}
% \centering
% %\begin{small}
% 			\begin{tabular}{lllll}
% 				\hline
% 				\hline
% 				Property/&Linear  &  iDeepViewLearn&	Randomized &Deep CCA$^{*}$,\\
%  								Methods& Methods& (Proposed) &	KCCA$^{*}$&Deep GCCA$^{+}$\\
% 		\hline
% 		\hline
% 		Nonlinear Relationships & &  \checkmark& \checkmark & \checkmark   \\
% 		\hline
% 		Classification& \checkmark& \checkmark&  &    \\
% 		\hline
% 		Variable ranking/selection& \checkmark&  \checkmark& &   \\
% 		\hline
% 		Covariates&\checkmark & \checkmark & &  \checkmark  \\
% 		\hline
% 	Smoothness&\checkmark & \checkmark&  &  \\
% 		\hline
% 		\hline
% \end{tabular}
% \caption{Unique features of iDeepViewLearn compared to other methods for associating data from multiple sources. *Only applicable to two views. +Covariates could be added as additional view in Deep GCCA.}
% %\end{small}
% \end{table}

\section{More on Simulations}
% \begin{figure}[H]
% \begin{tabular}{ccccc}
%          \centering
%          \includegraphics[width=0.18\textwidth]{Plots/deepIDAX1S2.png}& \includegraphics[width=0.18\textwidth]{Plots/deepIDAX2S2.png}& \includegraphics[width=0.18\textwidth]{Plots/deepIDAX1X2_15S2.png}& \includegraphics[width=0.18\textwidth]{Plots/deepIDAX1X2_18S2.png}& \includegraphics[width=0.18\textwidth]{Plots/deepIDAX1X2_10_5S2.png}\\
% \end{tabular}
%     \caption{Structure of nonlinear relationships between (First left panel) signal variables in View 1; (Second left panel) signal variables in View 2; (Middle panel)-(Fifth panel) signal variables between Views 1 and 2.}
%     \label{fig:nonlinears1}
% \end{figure}

% \begin{figure}[!htbp]
%     \centering
%     \includegraphics[width=0.3\linewidth]{Plots/scalefreenetwork2.png} \includegraphics[width=0.3\linewidth]{Plots/lattice2.png} \includegraphics[width=0.3\linewidth]{Plots/cluster3.png}\\
% %    (A) ~~~~& (B) ~~~~& C
%     \caption{Network structure for the first 50 variables in $\bX^{(1)}$ and $\bX^{(2)}$. Left: scale-free network; Middle: Lattice; Right: Cluster. For the Scale-free network, we consider variable 2 has a hub variable. Variable 2 and the variables directly connected to it are considered as signal variables. For the Lattice network, all variables except variable 50 are considered as signals. For the Cluster network, the circled clusters are considered as signals. }
%     \label{fig:network}
% \end{figure}
\subsection{Linear simulations when prior information is not available}
% ==================== LINEAR SIMULATION ====================

%\subsubsection{Linear Simulations}
We simulate data with $D=2$ views, and $K=3$ classes following the simulation setup in \cite{SIDA:2019}. For each view, we concatenate data from the three classes i.e.,  $\bX^{(d)} = [\bX_1^{(d)}, \bX_2^{(d)}, \bX_3^{(d)}], d=1,2$. The combined data $\left(\bX^{(1)}_k, \bX^{(2)}_k\right)$ for each class are simulated from $N(\bmu_k, \bSigma)$, where $\bmu_k = (\bmu^{(1)}_k, \bmu^{(2)}_k)^{\smt} \in \R^{p^{(1)} + p^{(2)}}, k=1,2,3$ is the combined mean vector for class $k$; $\bmu^{(1)}_k \in \R^{p^{(1)}}, \bmu^{(2)}_k \in \R^{p^{(2)}}$ are the mean vectors for $\bX^{(1)}_k$ and $\bX^{(2)}_k$ respectively. The true covariance matrix $\bSigma$ is partitioned as \begin{eqnarray}
\bSigma =\left(
\begin{array}{cc}
\bSigma^{1} & \bSigma^{12} \nonumber\\
\bSigma^{21} & \bSigma^{2} \nonumber\
\end{array} \right), \bSigma^1 =\left(
\begin{array}{cc}
\widetilde{\bSigma}^{1} & \textbf{0} \nonumber\\
\textbf{0} & \bI_{p-20} \nonumber\
\end{array} \right), \bSigma^2 =\left(
\begin{array}{cc}
\widetilde{\bSigma}^{2} & \textbf{0} \nonumber\\
\textbf{0} & \bI_{q-20} \nonumber\
\end{array} \right)
\end{eqnarray}
\noindent where $\bSigma^{1}$, $\bSigma^{2}$ are, respectively, the covariance of $\bX^{(1)}$ and $\bX^{(2)}$, and $\bSigma^{12}$ is the cross-covariance between the two views. $\widetilde{\bSigma}^{1}$ and $\widetilde{\bSigma}^{2}$ are each block diagonal with 2 blocks of size 10, between block correlation 0, and each block is a compound symmetric matrix with correlation 0.8. We generate $\bSigma^{12}$ as follows.  Let $\mathbf{V}^{1} = [\mathbf{V}^{1}_{1},~ \mathbf{0}_{(p^{(1)}-20) \times 2}]^{\smt} \in \R^{p^{(1)} \times 2 }$ where the entries of $V^{1}_{1} \in \R^{20 \times 2}$ are \textit{i.i.d} samples from U(0.5,1). We similarly define $\mathbf{V}^2$ for the second view, and normalize such that $\mathbf{V}^{1^{T}}\bSigma^{1}\mathbf{V}^{1} = \mathbf{I}$ and $\mathbf{V}^{2^{\smt}}\bSigma^2 \mathbf{V}^{2} = \mathbf{I}$. We then set $\bSigma^{12} = \bSigma^{1}\bV^1\bD\bV^{2^{\smt}} \bSigma^{2}$, $\bD= \text{diag}(\rho_1, \rho_2)$ to represent a moderate association between the views. For the separation between classes, we take $\bmu_k$ as the columns of $[\bSigma\bA, \textbf{0}_{p^{(1)} + p^{(2)}}]$, and $\bA=[\bA^1, \bA^2]^{\smt} \in \Re^{(p^{(1)} + p^{(2)}) \times 2}$. Here, the first column of $\bA^{1} \in \R^{p^{(1)} \times 2}$ is set to $(c_1\textbf{1}_{10}, \textbf{0}_{p^{(1)}-10})$ and
the second column is set to $( \textbf{0}_{10},-c\textbf{1}_{10}, \textbf{0}_{p^{(1)}-20})$. We set $\bA^{2} \in \R^{p^{(2)} \times 2}$ similarly. We consider different combinations of $(\rho_1, \rho_2,c)$. For each combination, we consider the equal class size $n_k=180$ and dimensions $(p^{(1)} /p^{(2)} = 1000/1000)$. The true number of signal variables is $20$.

The proposed method was implemented in the training data, and we identified the top 20 variables. We learned a new model with only the top 20 variables and used the learned model and the testing data to obtain the test error. We report the results of our method and the competitors in \cref{tab:Linear}. Compared to the nonlinear association-based method, Deep CCA, the proposed method achieved lower misclassification rates across all settings. The proposed method had comparable error rates in Settings 1 and 3, and lower error rate in Setting 2 compared to the linear association-based method. Compared to SVM in stacked data, the proposed method had lower average error rates across all settings. Since Deep CCA is not capable of variable selection, we coupled Deep CCA with the TS network. The performance of the proposed method in identifying the 20 signal variables was comparable with Sparse CCA and better than Deep CCA + TS. Furthermore, the proposed method had zero false positives, suggesting that the method was capable of not falsely ranking the noise variables in the top 20. 
%and ignoring the noise variables was superior when compared to Deep CCA + TS, and compara. 
The TS framework for ranking variables was suboptimal, as is evident from the TPR, FPR, and F measures for Deep CCA + TS. \textcolor{blue}{Random forest has very comparable classification and feature selection performance with our method in the simple linear simulations.} The detailed network structures of linear and nonlinear settings are shown in \cref{tab: DCCA}, respectively.

\begin{table}
\caption{Linear Settings: \textcolor{blue}{randomly select combinations of hyper-parameters to search over}. TPR-1; true positive rate for \textcolor{blue}{$\mathbf{X}^{(1)}$}. Similar for TPR-2. FPR; false positive rate for \textcolor{blue}{$\mathbf{X}^{(2)}$}. Similar for FPR-2; F-1 is the F measure for \textcolor{blue}{$\bX^{(1)}$}. Similar for F-2. The highest F-1/2 is in \textcolor{red}{red}. (The mean error of two views is reported for MOMA; \textcolor{blue}{MOMA + SVM means combining the feature selection part of MOMA and SVM.})}\label{tab:Linear}
%\label{t:linear}
\begin{center}
\resizebox{\textwidth}{!}{
\begin{tabular}{lrrrrrrr}
% \Hline
Method&Error (\%)&  	TPR-1&TPR-2 & FPR-1 &FPR-2  & F-1& F-2	\\
			\hline
            \hline
			
			    \textbf{Setting 1}& \\
			    $(\rho_1=0.9, \rho_2=0.7, c=0.5)$\\
			     %   iDeepViewLearn&0.10 (0.10)&  	100.00	&100.00	&0.00	&0.00	&100.00	&100.00\\
			        \textbf{iDeepViewLearn}&0.09 (0.07)&  	100.00	&100.00	&0.00	&0.00	&\textcolor{red}{100.00}	&\textcolor{red}{100.00}\\
                    \textcolor{blue}{iDeepViewLearn on stacked data}	&0.09 (0.09)& 100.00	&100.00	&0.00	&0.00	&\textcolor{red}{100.00}	&\textcolor{red}{100.00}\\
			        Sparse CCA + SVM & 0.10 (0.06) & 100.00 & 100.00 & 0.42 & 0.25 & 91.67 & 94.95 \\
                    Deep CCA + TS + SVM & 6.53 (1.77) & 3.50 & 2.25 & 1.97 & 1.99 & 3.50 & 2.25 \\
                    %RKCCA  &10.82&  	-	&-	&-	&-	&-	&-\\
                    MOMA & 25.18 (5.61) &  	86.50	&86.25	&0.28	&0.28	&86.50	&86.25\\
                    \textcolor{blue}{MOMA + SVM} & 0.04 (0.07) &  	86.50	&86.25	&0.28	&0.28	&86.50	&86.25\\
                    \textcolor{blue}{Random Forest on stacked data} & 0.02 (0.04) &  100.00	&100.00	&0.00	&0.00	&\textcolor{red}{100.00}	&\textcolor{red}{100.00}\\
                    SVM on stacked data & 0.16 (0.15) &  	-	&-	&-	&-	&-	&-\\
			    \hline
			    
			    \textbf{Setting 2}&\\
			    $(\rho_1=0.15, \rho_2=0.05, c=0.12)$\\
			     %   iDeepViewLearn&30.18 (1.63)&  100.00	&100.00	&0.00	&0.00	&100.00	&100.00\\
			        \textbf{iDeepViewLearn}&31.61 (1.21)&  	100.00	&100.00	&0.00	&0.00	&\textcolor{red}{100.00}	&\textcolor{red}{100.00}\\
                    \textcolor{blue}{iDeepViewLearn on stacked data}	&32.04 (1.40)& 100.00	&100.00	&0.00	&0.00	&\textcolor{red}{100.00}	&\textcolor{red}{100.00}\\
                    Sparse CCA + SVM & 38.60 (3.48) & 89.50 & 90.25 & 0.01 & 0.01 & 91.51 & 93.40 \\
			        Deep CCA + TS + SVM & 45.09 (2.02) & 2.50 & 2.25 & 1.99 & 1.99 & 2.50 & 2.25\\
			        %RKCCA  &59.99&  	-	&-	&-	&-	&-	&-\\
                    MOMA & 43.15 (2.49) &  	67.25	&67.75	&0.67	&0.66	&67.25	&67.75\\
                    \textcolor{blue}{MOMA + SVM} & 31.25 (2.30) &  	67.25	&67.75	&0.67	&0.66	&67.25	&67.75\\
                    \textcolor{blue}{Random Forest on stacked data}  & 31.20 (1.85) &  99.75	&100.00	&0.01	&0.00	&99.75	&\textcolor{red}{100.00}\\
			        SVM on stacked data  & 32.57 (1.89) &  	-	&-	&-	&-	&-	&-\\
			    \hline
			    
			    \textbf{Setting 3}&\\
			    $(\rho_1=0.9, \rho_2=0.7, c=0.5)$\\
			     %   iDeepViewLearn&0.01 (0.03)&  100.00	&100.00	&0.00	&0.00	&100.00	&100.00\\
			        \textbf{iDeepViewLearn}&0.01 (0.03)&  	100.00	&100.00	&0.00	&0.00	&\textcolor{red}{100.00}	&\textcolor{red}{100.00}\\
                    \textcolor{blue}{iDeepViewLearn on stacked data}	&0.04 (0.10)& 100.00	&100.00	&0.00	&0.00	&\textcolor{red}{100.00}	&\textcolor{red}{100.00}\\
                    Sparse CCA + SVM & 0.00 (0.00) & 100.00 & 100.00 & 0.31 & 0.24 & 93.71 & 94.83 \\
			        Deep CCA + TS + SVM & 5.63 (1.96) & 3.00 & 2.50 & 1.89 & 1.90 & 3.00 & 2.50\\
			        %RKCCA  &4.30&  	-	&-	&-	&-	&-	&-\\
                    MOMA & 19.26 (5.90) &  	94.25	&89.00	&0.12	&0.22	&94.25	&89.00\\
                    \textcolor{blue}{MOMA + SVM} & 0.00 (0.00) &  	94.25	&89.00	&0.12	&0.22	&94.25	&89.00\\
                    \textcolor{blue}{Random Forest on stacked data} & 0.00 (0.00) & 100.00	&100.00	&0.00	&0.00	&\textcolor{red}{100.00}	&\textcolor{red}{100.00}\\
			        SVM on stacked data & 0.03 (0.07) &  	-	&-	&-	&-	&-	&-\\
\hline
\hline
\end{tabular}}
\end{center}
\end{table}

\section{\textcolor{blue}{More on Experiment Settings}}

In this section, we present detailed hyper-parameter settings for our method and all competing methods to enhance reproducibility. The hyper-paramaters of our proposed method, MOMA (+ SVM), Deep CCA + SVM, Sparse CCA + SVM, SVM and random forest are shown in~\cref{tab: ours,tab: moma,tab: DCCA,tab: SCCA,tab: SVM,tab: rf}, respectively. The hyper-parameters of the methods that are trained with default parameters are listed in \cref{tab: other}.

\begin{table}
	\begin{small}
		\begin{centering}
			%	\begin{scriptsize}	
			\caption{iDeepViewLearn.}	\label{tab: ours}
   \resizebox{\textwidth}{!}{
			\begin{tabular}{llllll}
				\hline
				\hline
				Data & K  & lr & lrz & C & Gamma\\ 
				\hline
				\hline
				Linear Setting 1 & 16 & 0.0001 & 1 & 1 & scale\\
                Linear Setting 2 & 12 & 0.0001 & 1 & 0.1 & scale\\
                Linear Setting 3 & 12 & 0.0001 & 1 & 100 & scale\\
                Nonlinear Setting 1 & 10 & 0.0001 & 1 & 1 & scale\\
                Nonlinear Setting 2 & 10 & 0.0001 & 1 & 0.1 & scale\\
                Nonlinear Setting 3 & 40 & 0.0001 & 1 & 10 & scale\\
                Scale-free Setting 1 & 12 & 0.0001 & 10 & 1 & scale\\
                Scale-free Setting 2 & 16 & 0.1 & 1 & 1 & scale\\
                Lattice Setting 1 & 9 & 0.0001 & 1 & 10 & scale \\
                Lattice Setting 2 & 9 & 0.0001 & 1 & 10 & scale \\
                Cluster Setting 1 & 6 & 0.0001 & 10 & 1 & scale \\
                Cluster Setting 2 & 6 & 0.1 & 1 & 10 & scale\\
                Holm Breast Cancer Study (top $10\%$) & 30 & 0.1 & 1 & 0.1 & scale\\
                Holm Breast Cancer Study (top $20\%$) & 30 & 0 & 0.1 & 1 & scale\\
                LGG (top $50$) & 30 & 0.1 & 1 & 1 & scale\\
                LGG (top $100$) & 40 & 0.001 & 10 & 1 & scale\\
				\hline
				\hline
			\end{tabular}}
		\end{centering}
	\end{small}
\end{table}

\begin{table}
	\begin{small}
		\begin{centering}
			%	\begin{scriptsize}	
			\caption{MOMA and MOMA + SVM.}	\label{tab: moma}
   \resizebox{\textwidth}{!}{
			\begin{tabular}{llllll}
				\hline
				\hline
				Data & Number of modules  & Weight decay & Patience number & C & Gamma\\ 
				\hline
				\hline
				Linear Setting 1 & 64 & 0.001 & 50 & 10 & scale\\
                Linear Setting 2 & 128 & 0.001 & 100 & 0.1 & scale\\
                Linear Setting 3 & 128 & 0.001 & 100 & 0.1 & scale\\
                Nonlinear Setting 1 & 32 & 0.00 & 100 & 10 & scale\\
                Nonlinear Setting 2 & 128 & 0.001 & 50 & 1 & scale\\
                Nonlinear Setting 3 & 32 & 0.00 & 50 & 1 & scale\\
                Scale-free Setting 1 & 128 & 0.00 & 100 & 1 & scale\\
                Scale-free Setting 2 & 64 & 0.001 & 50 & 1 & scale\\
                Lattice Setting 1 & 128 & 0.00 & 100 &1 & scale \\
                Lattice Setting 2 & 32 & 0.00 & 50 & 1 & scale \\
                Cluster Setting 1 & 64 & 0.001 & 100 & 1 & scale \\
                Cluster Setting 2 & 128 & 0.00 & 50 & 10 & scale\\
                Holm Breast Cancer Study & 64 & 0.001 & 100 & 1 & scale\\
				\hline
				\hline
			\end{tabular}}
		\end{centering}
	\end{small}
\end{table}

\begin{table}
	\begin{small}
		\begin{centering}
			%	\begin{scriptsize}	
			\caption{Deep CCA+SVM, all with full batch for DCCA and rbf kernel for SVM.}	\label{tab: DCCA}
			\begin{tabular}{llllll}
				\hline
				\hline
				Data & Network Structure  & Epochs per run & C & Gamma\\
				\hline
				\hline
				Linear Setting 1 & Input-256*10-64-20 & 50 & 0.1 & scale \\
                Linear Setting 2  & Input-256*10-64-20 & 50 & 1  &  scale\\
                Linear Setting 3  & Input-256*10-64-20 & 50 &  0.1 & scale \\
                Nonlinear Setting 1  & Input-256*10-64-20 & 50 & 1 & scale \\
                Nonlinear Setting 2  & Input-256*10-64-20 & 50 & 10 & scale \\
                Nonlinear Setting 3  & Input-256*10-64-20 & 50 & 1  &  0.1\\
                Scale-free Setting 1 & Input-256*10-64-20 & 50 & 1 & scale\\
                Scale-free Setting 2 & Input-256*10-64-20 & 50 & 1 & scale\\
                Lattice Setting 1 & Input-256*10-64-20 & 50 & 1 & scale\\
                Lattice Setting 2 & Input-256*10-64-20 & 50 & 1 & scale\\
                Cluster Setting 1 & Input-256*10-64-20 & 50 & 1 & scale\\
                Cluster Setting 2 & Input-256*10-64-20 & 50 & 1 & scale\\
                % Graph simulations & Input-256*10-64-20 & 50 & 1 & scale\\
                Holm Breast Cancer Study & Input-256*10-64-20 & 50 & 1 & scale\\
                MNIST & Input-256*10-64-20 & 50 & 10 & scale\\
				\hline
				\hline
			\end{tabular}
		\end{centering}
	\end{small}
\end{table}	

\begin{table}
	\begin{small}
		\begin{centering}
			%	\begin{scriptsize}	
			\caption{Sparce CCA+SVM, all with ncancorr = 1, CovStructure = `Iden', nfolds = 5, ngrid = 10, standardize = TRUE, thresh = 0.0001, and maxiteration = 20 for SCCA and all with rbf kernel for SVM.}	\label{tab: SCCA}
			\begin{tabular}{lll}
				\hline
				\hline
				Data & C  & Gamma \\
				\hline
				\hline
				Linear Setting 1 & 0.1 & scale\\
                Linear Setting 2 & 0.1 & scale\\
                Linear Setting 3 & 1 & scale\\
                Nonlinear Setting 1 & 10 & 0.1\\
                Nonlinear Setting 2 & 1 & 1\\
                Nonlinear Setting 3 & 10 & 0.1\\
                Scale-free Setting 1 & 1 & scale\\
                Scale-free Setting 2 & 1 & scale\\
                Lattice Setting 1 & 1 & scale\\
                Lattice Setting 2 & 1 & scale\\
                Cluster Setting 1 & 1 & scale\\
                Cluster Setting 2 & 1 & scale\\
                % Graph simulations & 1 & scale \\
                Holm Breast Cancer Study & 1 & scale\\
				\hline
				\hline
			\end{tabular}
		\end{centering}
	\end{small}
\end{table}

\begin{table}
	\begin{small}
		\begin{centering}
			%	\begin{scriptsize}	
			\caption{SVM on stacked data, all with rbf kernel.}	\label{tab: SVM}
			\begin{tabular}{lll}
				\hline
				\hline
				Data & C  & Gamma \\
				\hline
				\hline
				Linear Setting 1 & 10  & scale\\
                Linear Setting 2 & 1 & scale\\
                Linear Setting 3 & 0.1 & scale\\
                Nonlinear Setting 1 & 1 & scale\\
                Nonlinear Setting 2 & 10 & scale\\
                Nonlinear Setting 3 & 1 & scale\\
                Scale-free Setting 1 & 1 & scale\\
                Scale-free Setting 2 & 1 & scale\\
                Lattice Setting 1 & 1 & scale\\
                Lattice Setting 2 & 1 & scale\\
                Cluster Setting 1 & 1 & scale\\
                Cluster Setting 2 & 1 & scale\\
                % Graph simulations & 1 & scale\\
                Holm Breast Cancer Study & 1 & scale\\
                MNIST & 1 & scale\\ % probably no tune
                LGG & 1 & scale \\
				\hline
				\hline
			\end{tabular}
		\end{centering}
	\end{small}
\end{table}	

\begin{table}
	\begin{small}
		\begin{centering}
			%	\begin{scriptsize}	
			\caption{Random forest.}	\label{tab: rf}
   \resizebox{\textwidth}{!}{
			\begin{tabular}{lllll}
				\hline
				\hline
				Data & n\_estimators  & max\_depth & min\_samples\_split & min\_samples\_leaf\\ 
				\hline
				\hline
				Linear Setting 1 & 200 & 20 & 5 & 2\\
                Linear Setting 2 & 200& None& 5& 4\\
                Linear Setting 3 & 200 & None & 5 & 1\\
                Nonlinear Setting 1 & 100 & None & 5 & 1\\
                Nonlinear Setting 2 & 200 & 30 & 2 & 1\\
                Nonlinear Setting 3 & 200 & None & 10 & 1 \\
                Scale-free Setting 1 & 200 & 30 & 2 & 1\\
                Scale-free Setting 2 & 200 & 20 & 5 & 2\\
                Lattice Setting 1 & 200 & None & 5 &1 \\
                Lattice Setting 2 & 200 & None & 10 & 2 \\
                Cluster Setting 1 & 50 & 20 & 10 & 4 \\
                Cluster Setting 2 & 200 & 30 & 2 & 1\\
                Holm Breast Cancer Study & 50 & 30 & 5 & 1\\
                LGG & 100 & 20 & 5 & 4\\
				\hline
				\hline
			\end{tabular}}
		\end{centering}
	\end{small}
\end{table}

\begin{table}
	\begin{small}
		\begin{centering}
			%	\begin{scriptsize}	
			\caption{Other methods that are trained with their default hyper-parameters.}	\label{tab: other}
			\begin{tabular}{lll}
				\hline
				\hline
				Method & Analysis & Default hyper-parameters \\
				\hline
				\hline
				Fused CCA & Simulation with  & method=`Fused',constopt=`OptB', mygamma=2, \\ 
                    & variable-variable connections & myeta=0.5, nfolds=5, ngrid=10, thresh=1e-04,\\
                    & & asnormalize=TRUE, maxiteration=20\\
                    & & and SVM with rbf kernel, C=1, gamma=`scale'\\
                    \hline
                    SIDA & LGG & gridMethod=`RandomSearch', AssignClassMethod=`Joint',\\
                    & & nfolds=5, ngrid=8, standardize=TRUE, maxiteration=20,\\
                    & & weight=0.5, thresh=1e-03\\
				\hline
                    Deep IDA & LGG & network structure=input-256*10-64-20, \\
                    & & learning rate=0.01, epoch=30 \\
                    \hline
                    DGCCA + SVM & LGG & network structure=input-128-128-20, \\
                    & & epochs=400, C=1, gamma=`scale', weight decay=0.0001\\
				\hline
			\end{tabular}
		\end{centering}
	\end{small}
\end{table}	

\clearpage
\bibliography{sn-bibliography}% common bib file
%% if required, the content of .bbl file can be included here once bbl is generated
%%\input sn-article.bbl